\documentclass{article}

\usepackage[preprint, nonatbib]{neurips_2019}

\usepackage[utf8]{inputenc} % allow utf-8 input
\usepackage[T1]{fontenc}    % use 8-bit T1 fonts
\usepackage{hyperref}       % hyperlinks
\usepackage{url}            % simple URL typesetting
\usepackage{booktabs}       % professional-quality tables
\usepackage{amsfonts}       % blackboard math symbols
\usepackage{nicefrac}       % compact symbols for 1/2, etc.
\usepackage{microtype}      % microtypography

\usepackage{amsmath}
\usepackage{amsfonts}
\usepackage{amssymb}
\usepackage{amsthm}
\usepackage{upgreek}

\usepackage{graphicx}
\usepackage{float}

\newcommand{\bE}{\mathbb{E}}
\newcommand{\N}{\mathcal{N}}

\title{Mutual Information Scaling and Expressive Power of Sequence Models}

\author{
	Huitao Shen\\
	Department of Physics\\
	Massachusetts Institute of Technology\\
	\texttt{huitao@mit.edu} \\
}

\begin{document}

\maketitle

\begin{abstract}
	Sequence models assign probabilities to variable-length sequences such as natural language texts. The ability of sequence models to capture temporal dependence can be characterized by the temporal scaling of correlation and mutual information. In this paper, we study the mutual information of recurrent neural networks (RNNs) including long short-term memories and self-attention networks such as Transformers. Through a combination of theoretical study of linear RNNs and empirical study of nonlinear RNNs, we find their mutual information decays exponentially in temporal distance. On the other hand, Transformers can capture long-range mutual information more efficiently, making them preferable in modeling sequences with slow power-law mutual information, such as natural languages and stock prices. We discuss the connection of these results with statistical mechanics. We also point out the non-uniformity problem in many natural language datasets. We hope this work provides a new perspective in understanding the expressive power of sequence models and shed new light on improving the architecture of them. 
\end{abstract}

\section{Introduction}
Recent years have witnessed the success of sequence machine learning models in tasks such as speech recognition \cite{6638947}, machine translation \cite{NIPS2014_5346,Bahdanau2014,NIPS2017_7181}, text summarization \cite{Rush2015,K16-1028} and music generation \cite{Eck2002}. In particular, generative sequence modeling is usually framed as an unsupervised learning problem, that is the estimation of the joint probability distribution of variable-length sequences $\mathbf{x}=(x_1,\ldots,x_n)$. For example, in word-level language modeling, $\mathbf{x}$ is usually a sentence and $x_i$ is the $i$-th word in the sentence \cite{Jelinek1980,Bengio2003}. In image modeling, $\mathbf{x}$ is the image and $x_i$ is the value of the $i$-th pixel in  the image \cite{Oord2016}. At the core of most sequence models is the factorization of the joint distribution to conditional distributions:
\begin{equation}
p(x_1,\ldots,x_t)=\prod_{i=1}^t p(x_i|x_1,\ldots,x_{i-1}). \label{eq:jc}
\end{equation}
For instance, in $n$-gram model the conditional distribution only depends on the recent $n$ elements in the sequence. In recurrent neural networks (RNNs), the conditional distribution implicitly depends on the entire history of the sequence through hidden states represented by fixed-length vectors. In self-attention networks like Transformers \cite{NIPS2017_7181}, the conditional distribution explicitly depends on the entire history of the sequence. 

The way that the sequence history is exploited in the conditional distribution profoundly determines the temporal correlation in the joint probability distribution. It is obvious the $n$-gram model cannot capture dependence longer than $n$ time steps like unbounded syntactic movements. A formal mathematical statement of the above fact can be made through the temporal scaling property of the model: In the joint distribution generated by $n$-gram models, the mutual information of symbols decays exponentially in temporal distance. When the correlation or the mutual information between two symbols are small enough, the model cannot distinguish the dependence between them and the noise. Beyond the simple $n$-gram model, it is known that both regular language and hidden Markov models (HMMs) have similar exponential temporal scaling behavior \cite{Li1987,Lin2017}. 

On a seemingly separate note, there have been intense interests in the statistical analysis of natural sequences since the 1990s. It is found that the slow algebraic or power-law decay of mutual information is ubiquitous in natural sequences including human DNA sequences \cite{Li_1992,Peng1992,PhysRevLett.68.3805}, natural languages \cite{doi:10.1142/S0218348X93000083,Ebeling_1994,doi:10.1142/S0218348X02001257}, computer programs \cite{doi:10.1142/S0218348X93000083,Kokol1999}, music rhythms \cite{Levitin3716,6791788},  stock prices \cite{10.2307/2938368,DING199383}, etc.. The origin of the power-law scaling behavior is still debated and is not the focus of this paper. Nevertheless, it is clear the exponential temporal scaling in models such as $n$-gram models and HMMs sets a fundamental limitation on their ability to capture the long-range dependence in natural sequences. 

It is then natural to ask the question what the temporal scaling behavior is in sequence model architectures such as RNNs and self-attention networks and why. In this paper, we study the mutual information scaling of RNNs and Transformers. We show that the mutual information decays exponentially in temporal distance, rigorously in linear RNNs and empirically in nonlinear RNNs including long short-term memories (LSTMs) \cite{doi:10.1162/neco.1997.9.8.1735}. In contrast, long-range dependence, including the power-law decaying mutual information, can be captured efficiently by Transformers. This indicates Transformers are more suitable to model natural sequences with power-law long-range correlation. We also discuss the connection of these results with statistical mechanics. Finally, we notice there is discrepancy in the statistical property between training and validation sets in many natural language datasets. This non-uniformity problem may prevent sequence models from learning the long-range dependence.

\section{Related Work}
\paragraph{Expressive Power of RNNs} Essentially, this work studies the expressive power of RNNs and Transformers.  Closely related works are Refs.~\cite{NIPS2013_5166,Karpathy2015,P18-1027}, where different approaches or metrics are adopted to empirically study the ability of RNNs as language models to capture long-range temporal dependence. In Refs.~\cite{NIPS2013_5166,Karpathy2015}, character-level RNNs and LSTMs are shown to be able to correctly close parentheses or braces that are far apart. In Ref.~\cite{P18-1027}, ablation studies found word-level LSTMs have an effective context size of around 200 tokens, but  only sharply distinguishes the recent 50 tokens.

\paragraph{Mutual Information Diagnosis} Mutual information flow in the network training is studied in Refs.~\cite{Schwartz-ziv,Goldfeld}. Using temporal scaling property to diagnose sequence modeling is rarely mentioned in the context of deep learning. The only works the author is aware of are Refs.~\cite{Lin2017,10.1371/journal.pone.0189326}. In Ref.~\cite{Lin2017}, it is argued that in theory deep LSTMs are effective in capturing long-range correlations. Although it is a tempting proposal, shallow LSTMs are empirically found to actually perform as well as, if not better, than deep LSTMs \cite{Melis2017}. The empirical study in this work on multiple datasets also confirm the same result. Ref.~\cite{10.1371/journal.pone.0189326} is an empirical study of natural language models and focuses mainly on the ``one-point'' statistics like Zipf's law. In the last section, it is mentioned that LSTMs can not reproduce the power decay of the autocorrelation function in natural languages, which is consistent with the findings of this work. 

To the best of the author's knowledge, this work is the first that theoretically studies the mutual information scaling in RNNs. It is also the first work that systematically studies the mutual information scaling in RNNs and Transformers. 

\section{Mutual Information}
The mutual information between two random variables is defined as
\begin{equation}
I(X;Y)\equiv \bE_{(X,Y)\sim p_{XY}}\left[\ln \frac{p_{XY}(x,y)}{p_X(x)p_Y(y)}\right].
\end{equation}
It has many equivalent definitions such as $I(X;Y)=H(X)+H(Y)-H(X,Y)$, where $H(X)$ is the entropy of the random variable $X$. Roughly speaking, it measures the dependence between two random variables. Consider a discrete-time random process $\{x_t:t\in\mathbb{N}\}$. With the above definition of mutual information, the auto-mutual information of the random process is defined as $I_x(i,j)\equiv I(x_i;x_j)$. The random process is assumed to be stationary such that the auto-mutual information only depends on the temporal distance between the random variables. In this case, auto-mutual information can be characterized solely by the time lag $\tau$: $I_x(\tau)\equiv I(x_t;x_{t+\tau})$. In the rest of this paper, we always assume the stationarity and adopt the above definition. We also use ``mutual information'' and ``auto-mutual information'' interchangeably, and drop the subscript $x$ in $I_x$ when the underlying random variable is evident. 

At least two notions of ``expressive power''  can be defined from the auto-mutual information: (i) the scaling behavior with the temporal distance, e.g. whether $I(\tau)$ decays algebraically or exponentially with $\tau$; (ii) the absolute magnitude of the mutual information $I(\tau)$ for a given $\tau$. In this paper, we will mainly focus on the first notion when $\tau$ is large, which ideally is determined by the intrinsic structure of the model. The second notion is not as universal and is also critically affected by the number of parameters in the model.

\section{Recurrent Neural Networks}
\subsection{Linear RNNs as Gaussian Processes}
We start with an analytical analysis of mutual information scaling in linear RNNs with Gaussian output. Consider the classical Elman RNN with the linear activation:
\begin{align}
h_t=&W_h x_{t-1} + U_h h_{t-1}, \label{eq:ht}\\
o_t=&U_o h_{t-1}. \label{eq:ot}
\end{align}
$o_t\in\mathbb{R}^d$ parameterizes the probability distribution from which $x_t$ is sampled, and $h_t\in\mathbb{R}^m$ is the hidden state. In the following, we assume $x_t$ is sampled from a multivariate Gaussian distribution with mean $o_t$ and covariance matrix proportional to the identity, i.e. $p(x_t|x_{t-1},\ldots,x_0,h_0)\sim \N(o_t,\sigma^2 I_d)$. It follows from iteratively applying Equation \eqref{eq:ht} that
\begin{equation}
o_t=U_oU_h^{t}h_0+\sum_{i=0}^{t-2} U_o U_h^{t-1-i} W_h x_i, \label{eq:his}
\end{equation}
Since $o_t$ depends on the entire history of $x_i$, $i=0,\ldots,t-1$,  the random process specified by $\{x_t:t\in\mathbb{N}\}$ is not Markovian. Therefore, it is not obvious how mutual information decays with temporal distance. In the following, we sketch the proof that the mutual information $I_x(\tau)$ in the above RNN decreases exponentially with time $\tau$ if the RNN does not simply memorize the initial condition. The full proof is presented in Appendix. 

The hidden state is often initialized as $h_0=0$. Under this initial condition, $p(x_0)$ is multivariate Gaussian, and so does the joint distribution of the entire sequence  $p(x_t,\ldots,x_0,h_0)$. In this way, the random process is a Gaussian process. Since we are interested in the mutual information between $x_{t_0}$ and $x_{t_0+t}$ for some generic $t_0$, without loss of generality we can set $t_0=0$ and let $p(h_0,x_0)$ be a generic multivariate Gaussian distribution. We can also let $p(h_0,x_0)$ be the distribution that is already averaged over the entire sequence. In any case, we will see the asymptotic behavior of the mutual information is almost independent of $p(h_0,x_0)$. 

We are interested in the distribution $p(x_0,x_t)$, hence block covariance matrices $\Sigma_{x_tx_0}$ and $\Sigma_{x_tx_t}$. $\Sigma_{x_tx_0}$ can be derived recursively as
\begin{equation}
\Sigma_{x_tx_0}=U_oU_h^{t}\Sigma_{h_0x_0} +\sum_{i=0}^{t-2}U_oU_h^{t-1-i}W_h\Sigma_{x_ix_0},
\end{equation}
which can be solved with generating functions. Define the formal power series $\Sigma(z)\equiv\sum_{n=0}^{\infty}\Sigma_{x_nx_0} z^n$. Its closed-form expression is computed as
\begin{align}
\Sigma(z)=\left[I_d-U_o(I_m-U_h z)^{-1}U_h W_h z^2\right]^{-1}\left[U_o(I_m-U_h z)^{-1} U_h\Sigma_{h_0x_0} z+\Sigma_{x_0x_0}\right]. \label{eq:sigx}
\end{align}
The long time asymptotic behavior of $\Sigma_{x_tx_0}$ can be analyzed by treating $\Sigma(z)$ as a function on the complex plane and studying its singularities \cite{flajolet2009analytic}. Because Equation \eqref{eq:sigx} is a rational function of $z$, it can be shown that elements in $\Sigma_{x_tx_0}$ either decrease or increase exponentially with $t$, and the exponent is bounded by only $W_h$, $U_h$ and $U_o$, independent of the initial condition $p(x_0,h_0)$. In the exponentially increasing case, $\Sigma_{x_tx_0}$ simply remembers the initial condition, which is not desirable. Therefore, in any working network every element of $\Sigma_{x_tx_0}$ decreases exponentially in $t$. 

The mutual information between $x_0$ and $x_t$ is computed as
\begin{equation}
I(x_0;x_t)
=-\frac{1}{2}\ln \det(I_d-\Sigma_{x_0x_0}^{-1}\Sigma_{x_0x_t}\Sigma_{x_tx_t}^{-1}\Sigma_{x_0x_t}^T)
\approx\frac{1}{2}\mathrm{tr}(\Sigma_{x_0x_0}^{-1}\Sigma_{x_0x_t}\Sigma_{x_tx_t}^{-1}\Sigma_{x_0x_t}^T). \label{eq:tr}
\end{equation}
$\Sigma_{x_0x_0}$ is time-independent. $\Sigma_{x_tx_t}$ can be proved to be non-degenerate in the $t\to\infty$ limit, because $x_t$ is sampled conditionally from a Gaussian distribution with covariance matrix $\sigma^2I_d$.  Therefore, $\Sigma_{x_tx_t}^{-1}$ tends to a finite constant matrix when $t$ is large. Each element in $\Sigma_{x_0x_t}$ decays exponentially with $t$. 
In this way, elements in $\Sigma_{x_0x_0}^{-1}\Sigma_{x_0x_t}\Sigma_{x_tx_t}^{-1}\Sigma_{x_0x_t}^T$ is exponentially small when increasing $t$, which justifies the last equality in Equation \eqref{eq:tr}. Because trace is a linear function, the mutual information also decreases exponentially with $t$. This finishes the proof that in any linear Elman RNN with Gaussian output that does not simply memorize the initial condition, the mutual information decays exponentially with time. We note that adding bias terms in Equation \eqref{eq:ht} and \eqref{eq:ot} does not affect the conclusion because the mutual information of Gaussian random variables only depends on the covariance matrix, while the bias terms only affect their mean. We will talk more about this result at the Discussion section. 

\subsection{Nonlinear RNNs}
\subsubsection{Binary Sequence}
We now study how the linear RNN result generalizes to nonlinear RNNs on symbolic sequences. Our first dataset is artificial binary sequences. This dataset is simple and clean, in the sense that it only contains two symbols and is strictly scaleless. The training set contains 10000 sequences of length 512, whose mutual information decays as $I(\tau)=0.1\tau^{-0.4}$. During training, we do not truncate the backpropagation through time (BPTT). After training, we unconditionally generate 2000 sequences and estimate their mutual information. The generation algorithm of the dataset, along with experiment details, is reported in Appendix. 

\begin{figure}
	\centering
	\includegraphics[width=0.45\textwidth]{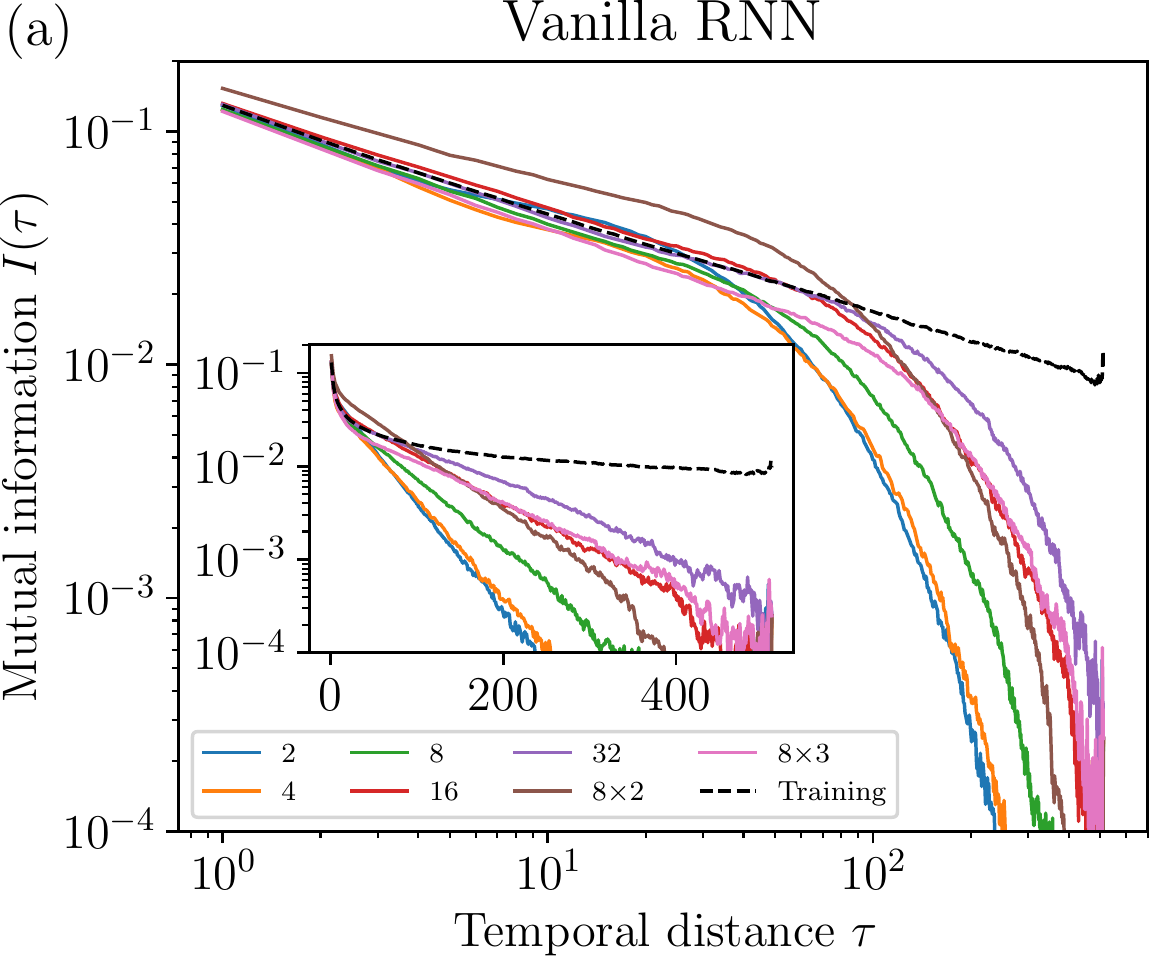}
	\hspace{4pt}
	\includegraphics[width=0.45\textwidth]{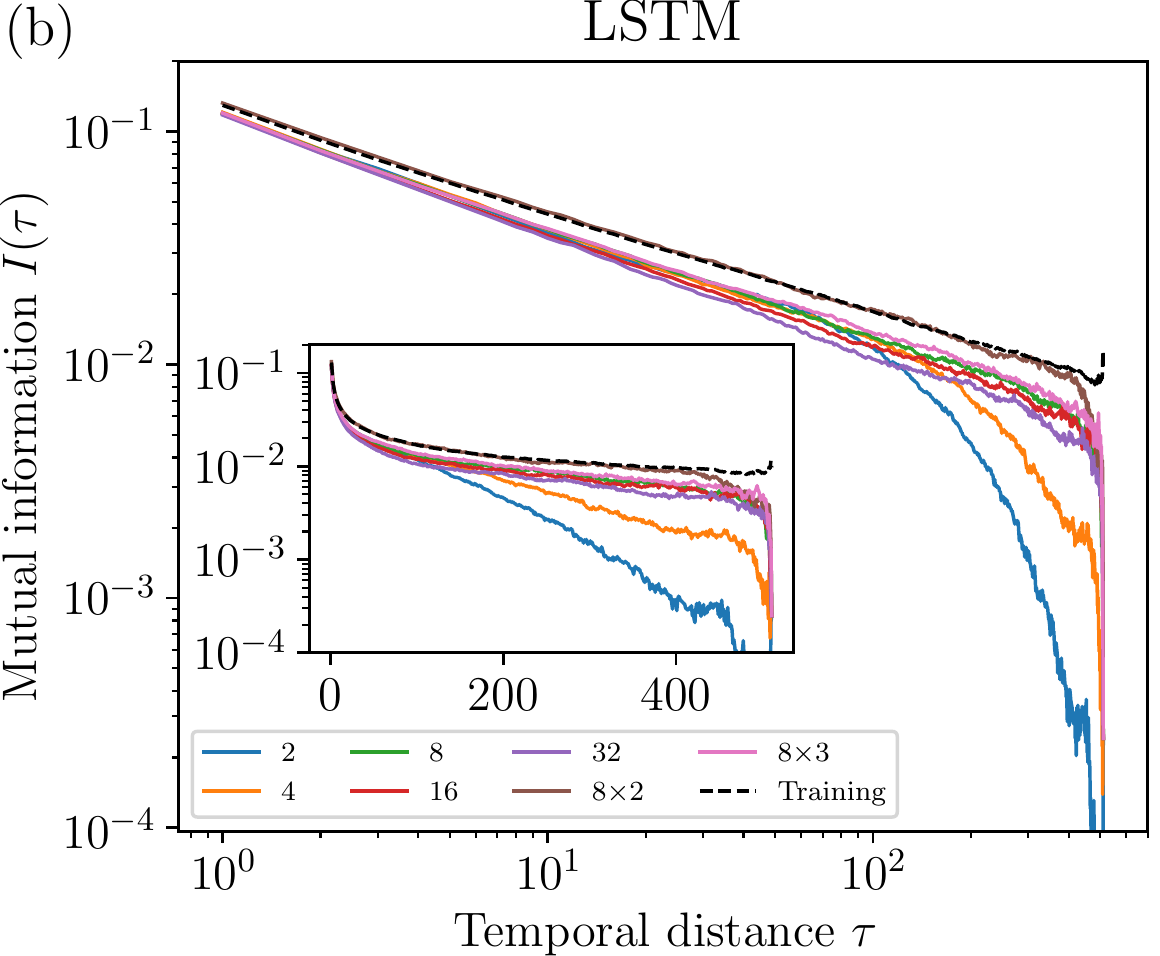}
	\vspace{8pt}
	
	\includegraphics[width=0.45\textwidth]{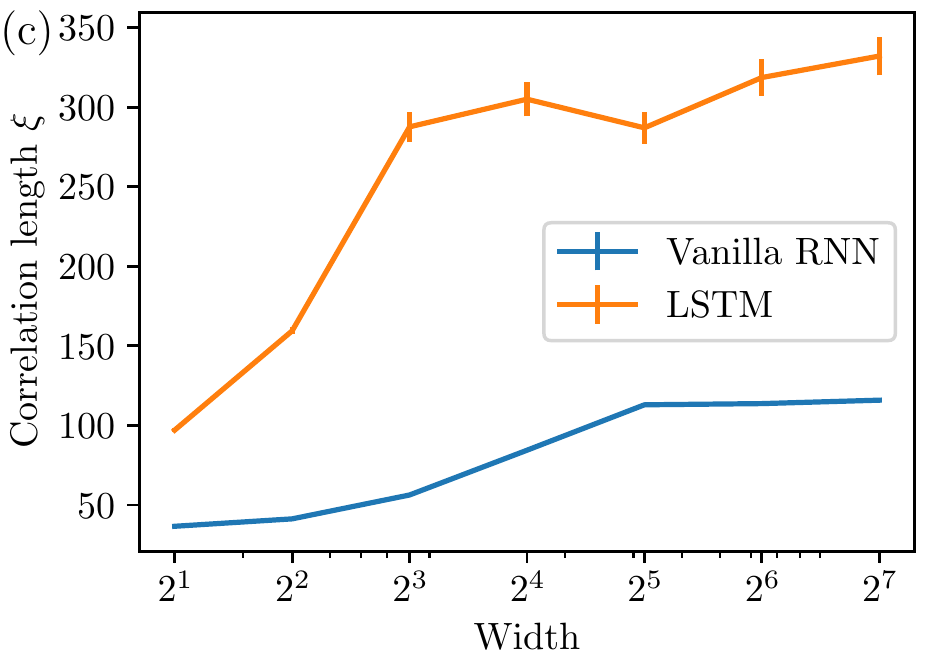}
	\hspace{4pt}
	\includegraphics[width=0.45\textwidth]{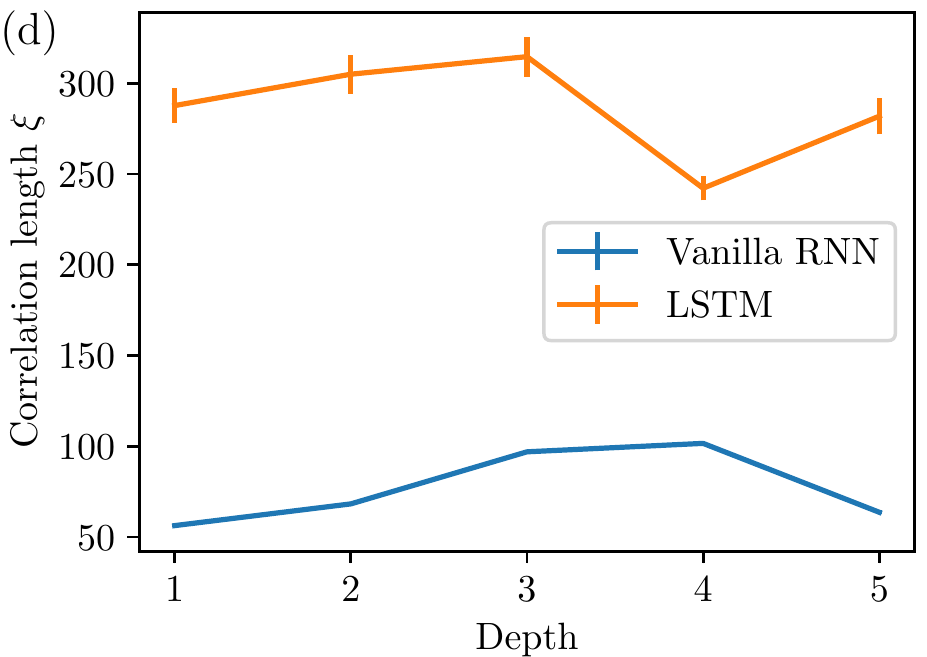}
	\caption{(a) Estimated mutual information of unconditionally generated sequences from vanilla RNNs on binary sequences. The legend denotes width($\times$depth, if depth$>$1); (b) Same as (a) but for LSTMs; (c) Fitted correlation length $\xi$ as a function of the RNN width. The depth of all networks is one. Only data points greater than $10^{-3}$ is used due to the estimation error in the long distance. The error bar represents the 95\% confidence interval. (d) Same as (c) but as a function of the RNN depth. The width of all networks is 8. }
	\label{fig:binrnn}
\end{figure}

\paragraph{Vanilla RNN} It is very clear from the straight line in the semi-log plot (inset of Figure \ref{fig:binrnn}(a)) that $I(\tau)$ decays exponentially with $\tau$ in vanilla RNNs:
\begin{equation}
	I(\tau)=I_0 e^{-\tau/\xi},
\end{equation}
where we have defined the ``correlation length'' $\xi$. 

If $\xi$ increases very rapidly with the width (hidden unit dimension of RNN layers) or the depth (number of RNN layers) of the network, the exponential decay will not bother us practically. However, this is not the case. The correlation length as a function of the network width is fitted in Figure \ref{fig:binrnn}(c).  For small networks ($m\leq 32$), the correlation length increases logarithmically with the hidden unit dimension $\xi\sim \ln m$. When the network becomes large enough, the correlation length saturates to around $\xi\sim 100$. The almost invisible error bars suggest the goodness of the exponential fitting of the temporal mutual information. The correlation length as a function of the network depth is fitted in Figure \ref{fig:binrnn}(d), which increases linearly for shallow networks. Therefore, increasing the depth of the vanilla RNNs is more efficient in capturing the long-range temporal correlation then increasing the width. For relatively deep networks, the performance deteriorates probably due to the increased difficulty in training. Interestingly, the $8\times 2$ network in Figure \ref{fig:binrnn}(a) overestimates the short-range correlation in order to compensate the rapid exponential decay in the long distance. 

\paragraph{LSTM} LSTMs perform much better in capturing long-range correlations, although the exponential decay can still been seen clearly in very small LSTMs (Figure \ref{fig:binrnn}(b) inset). When $m>8$, it is hard to distinguish the exponential decay from the algebraic decay. Nevertheless, we still fit the correlation length. Note that the fitting on the training set yields the baseline $\xi\approx 420$, which is comparable to the sequence length. The correlation length also increases linearly with width in small LSTMs and then saturates. The depth dependence is similar to that of vanilla RNNs too, which is consistent with previous studies that shallow LSTMs usually perform as well as, if not better, than deep LSTMs \cite{Melis2017}. 

To summarize, both vanilla RNNs and LSTMs show exponential decaying mutual information on the binary dataset. The correlation length $\xi$ of LSTMs has a much better scaling behavior than vanilla RNNs.

\subsubsection{Natural Language}
We extend the scaling analysis to the more realistic natural language dataset WikiText-2 \cite{Merity2016}. Since vanilla RNNs cannot even capture the long-range dependence in the simple binary dataset, we only focus on LSTMs here. During training, the BPTT is truncated to 100 characters at character level and 50 tokens at word level. After training, we unconditionally generate a long sequence with 2MB characters and estimate its mutual information \emph{at character level}. 

Different from binary sequences, there exists multiple time scales in WikiText-2. In the short distance $\tau\lesssim 10$  (word level), the mutual information decays exponentially, potentially due to the arbitrariness of characters within a word. In the long distance $50\lesssim \tau\lesssim 1000$ (paragraph level), the mutual information follows a power-law decay $I(\tau)\sim \tau^{-0.3}$. The qualitative behavior is mixed at the intermediate distance (sentence level). 

Strikingly, there is significant discrepancy between the mutual information profile between the training and the validation set. Not only the mutual information on the validation set is much larger, the algebraic decay in the long distance is missing as well. The fact that we always pick the best model on the validation set for sequence generation may prevent the model from learning the long-range mutual information. Therefore, we should interpret results especially from word-level models cautiously. We also find similar non-uniformity in datasets such as Penn Treebank. See Appendix for more details. 

\begin{figure}
	\centering
	\includegraphics[width=0.45\textwidth]{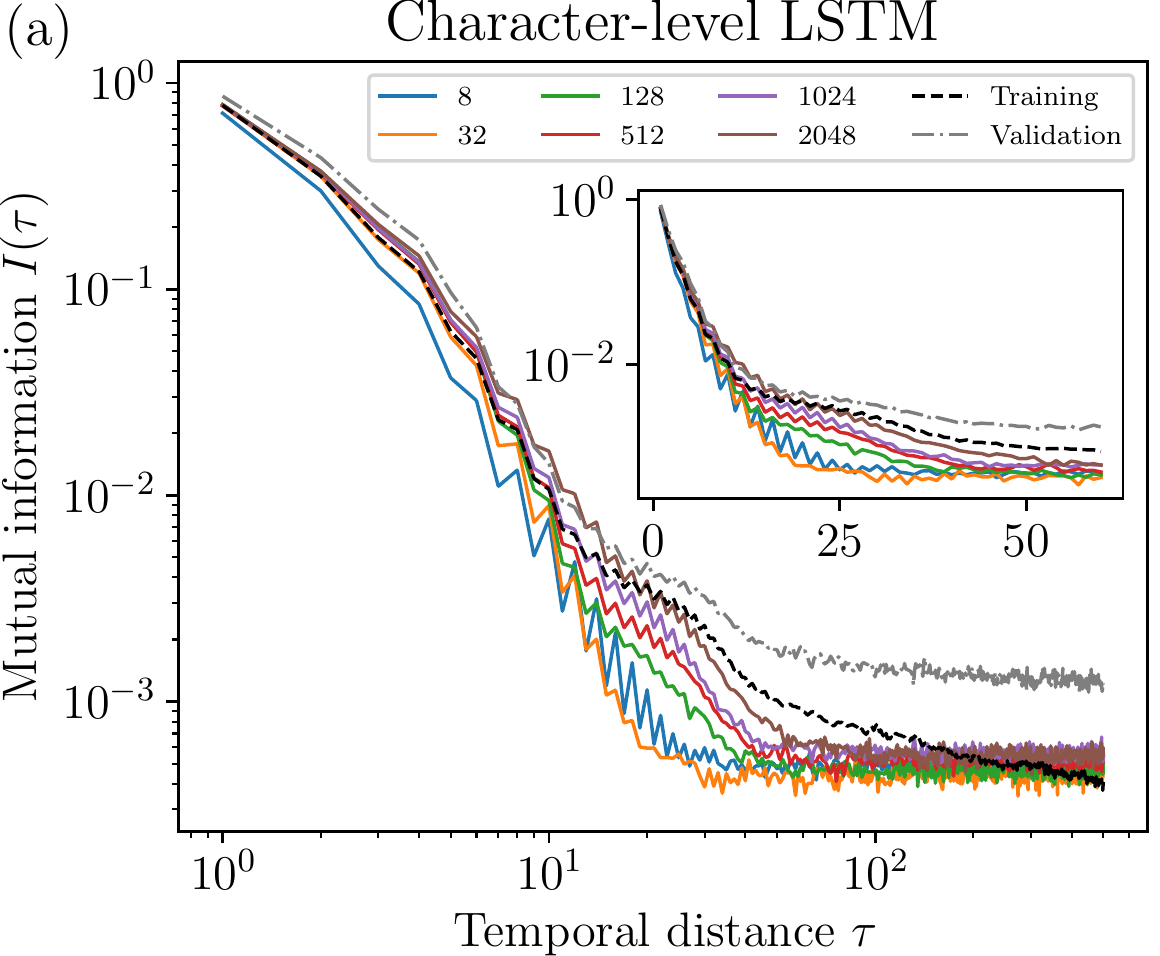}
	\hspace{4pt}
	\includegraphics[width=0.455\textwidth]{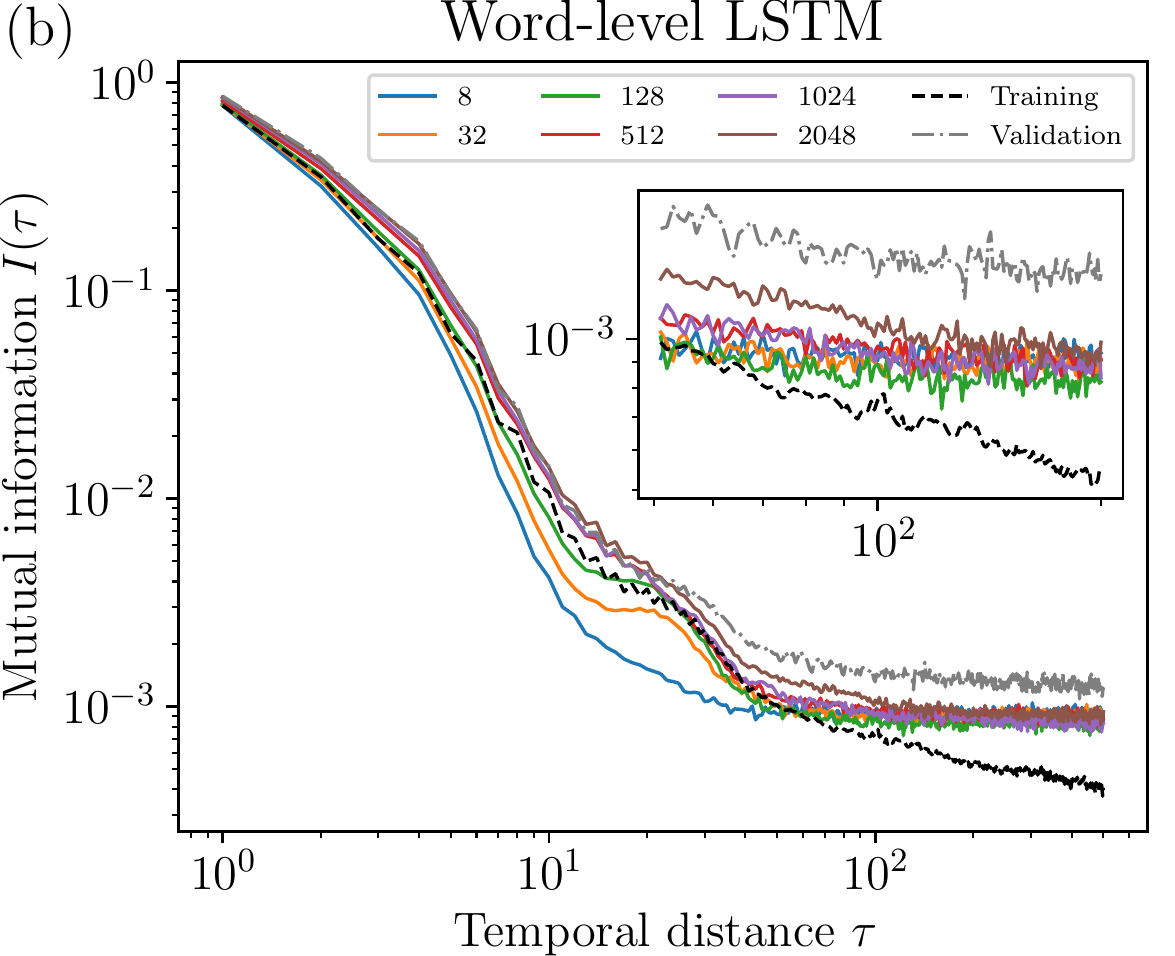}
	\caption{Estimated mutual information of unconditionally generated sequences from LSTMs on WikiText-2. The legend denotes width. The depth of all networks is one. (a) Character-level LSTM; (b) Word-level LSTM. }
	\label{fig:textlstm}
\end{figure}

Character-level LSTMs can capture the short-range correlation quite well as long as the width is not too small ($m\geq 32$). In the intermediate distance, it seems from the inset of Figure \ref{fig:textlstm}(a) that all LSTMs show an exponential decaying mutual information. In large models where $m=1024,2048$, the short-range mutual information is overestimated to compensate the rapid decay in the intermediate distance. No power-law decay in the long distance is captured at all. 

Word-level LSTMs can capture the mutual information up to the intermediate distance. There is no surprise the short-range correlation is captured well as the word spelling is already encoded through word embedding. There is even long-range power-law dependence up to $\tau\lesssim 150$ in the largest model with $m=2048$, although the mutual is overestimated approximately two times throughout all distances. 

In this dataset, the mutual information of LSTMs always decays to some nonzero constant instead of to zero like that in binary sequences. We speculate that this is likely due to the short-range memory effect, similar to the non-decaying mutual information in repetitive sequences. The simplest example is the binary sequence where 01 and 10 are repeated with probability $p$ and $1-p$ respectively, which can be generated by a periodic HMM. It is not hard to prove the mutual information is a constant for all temporal distances greater than one. See Appendix for a proof.

\section{Self-Attention Networks}
We now turn to the empirical study of the original Transformer model \cite{NIPS2017_7181}. In principle, the conditional distribution in Transformers explicitly depends on the entire sequence history. For complexity reasons, during the training the look-back history is usually truncated to the recent $n$ elements. In this sense, Transformers are like $n$-gram models with large $n$. For the purpose of this paper, the truncation will not bother us, because we are only interested in the mutual information with $\tau\lesssim n$. On WikiText-2, $n$ is limited to 512 characters at character level and 128 tokens at word level. However, if one is really interested in $\tau\gg n$, the result on the $n$-gram model suggests that the mutual information is bound to decay exponentially. The network width, which is the total hidden unit dimension, is defined as $m=\text{number  of heads}\times\text{hidden dimension of each head}$. For binary sequences, we use Transformers with four heads and for WikiText-2, we use eight heads. 

\subsection{Binary Sequence}
Transformers can very efficiently capture long-range dependence in binary sequences. A single-layer Transformer of width $m=16$ can already capture the algebraic behavior quite well. Interestingly, the mutual information does not decay exponentially even in the simplest $m=4$ model. Moreover, the magnitude of the mutual information always coincides with the training data very well, while that of LSTMs is almost always slightly smaller. The correlation length fitting can be found in Appendix, although the mutual information does not fit quite well with the exponential function. 

\begin{figure}
	\centering
	\includegraphics[width=0.45\textwidth]{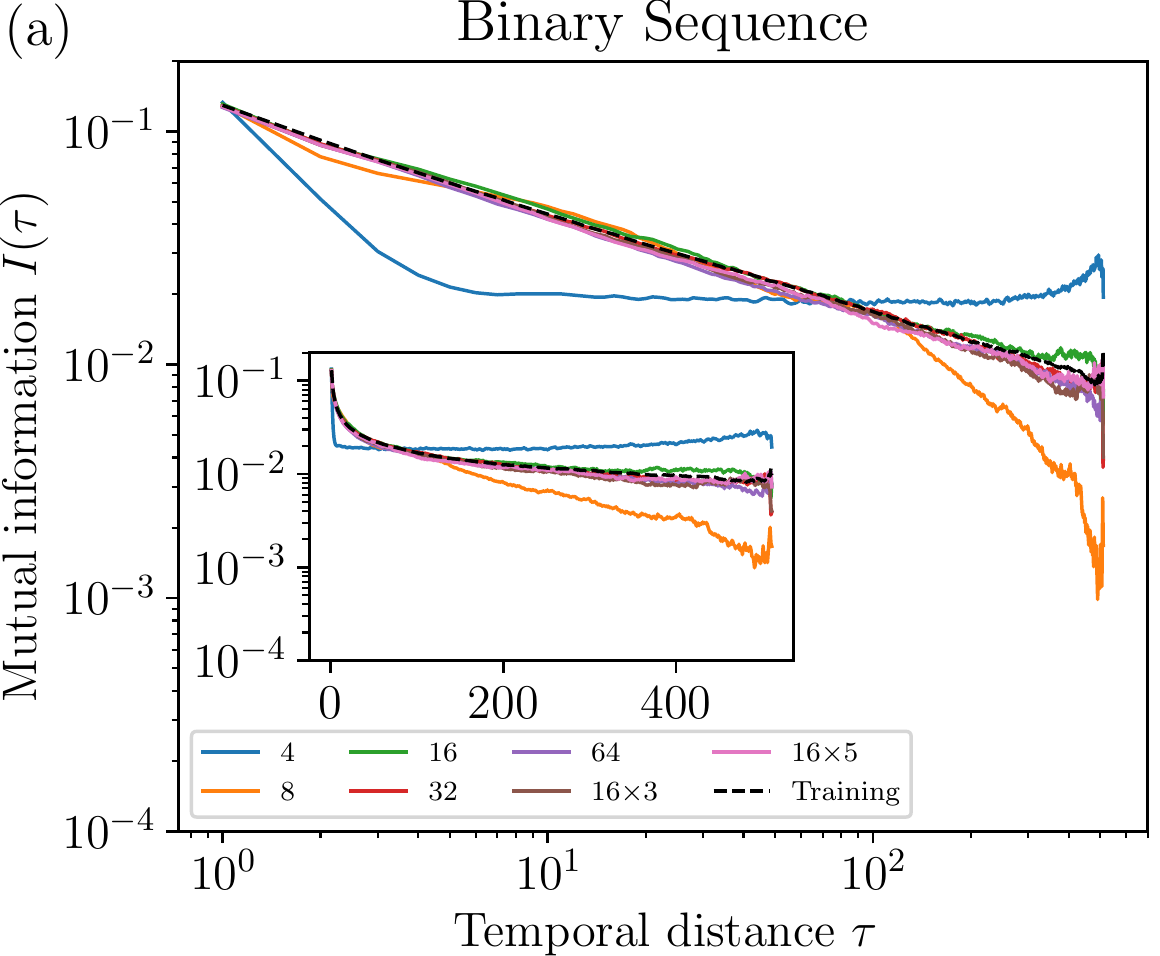}
	\hspace{4pt}
	\includegraphics[width=0.455\textwidth]{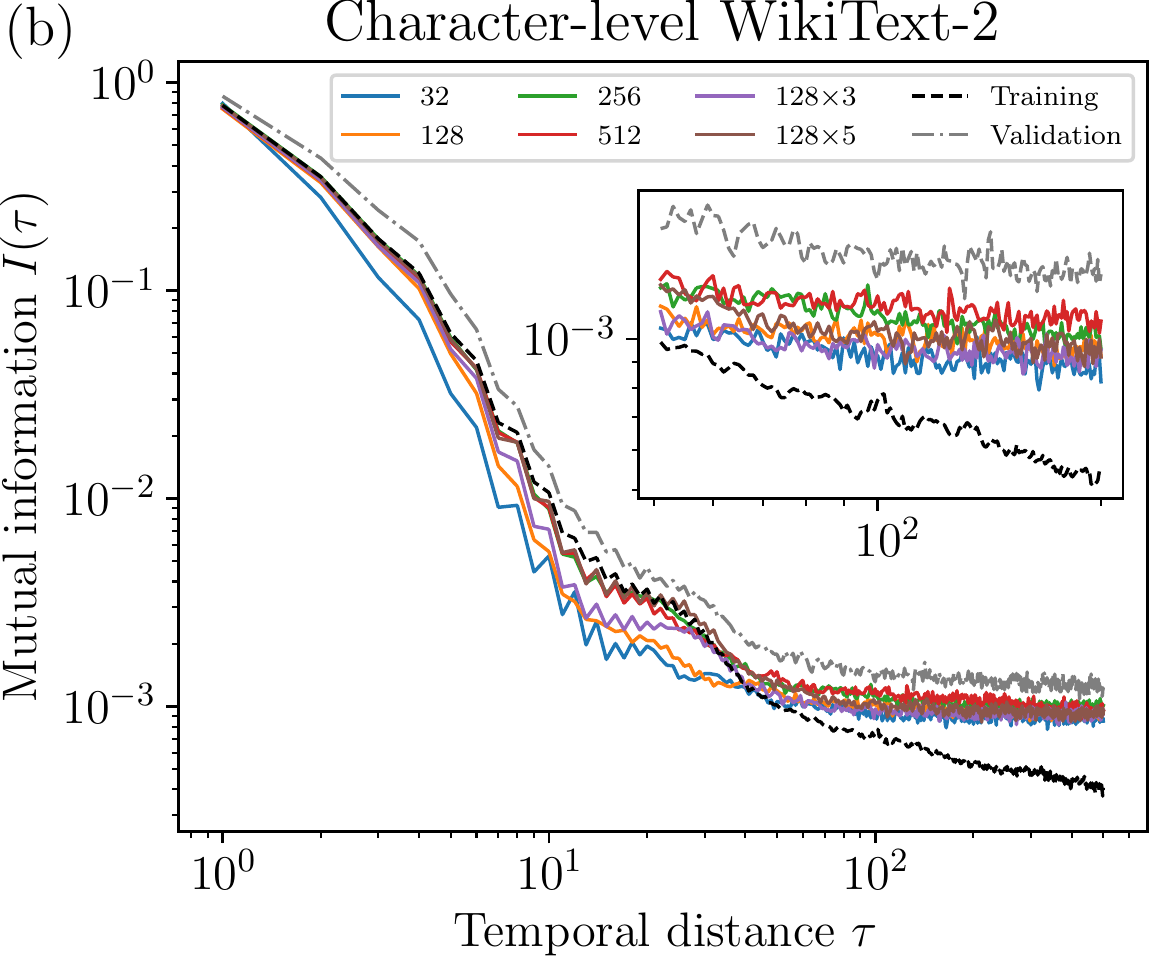}
	\vspace{8pt}
	
	\includegraphics[width=0.455\textwidth]{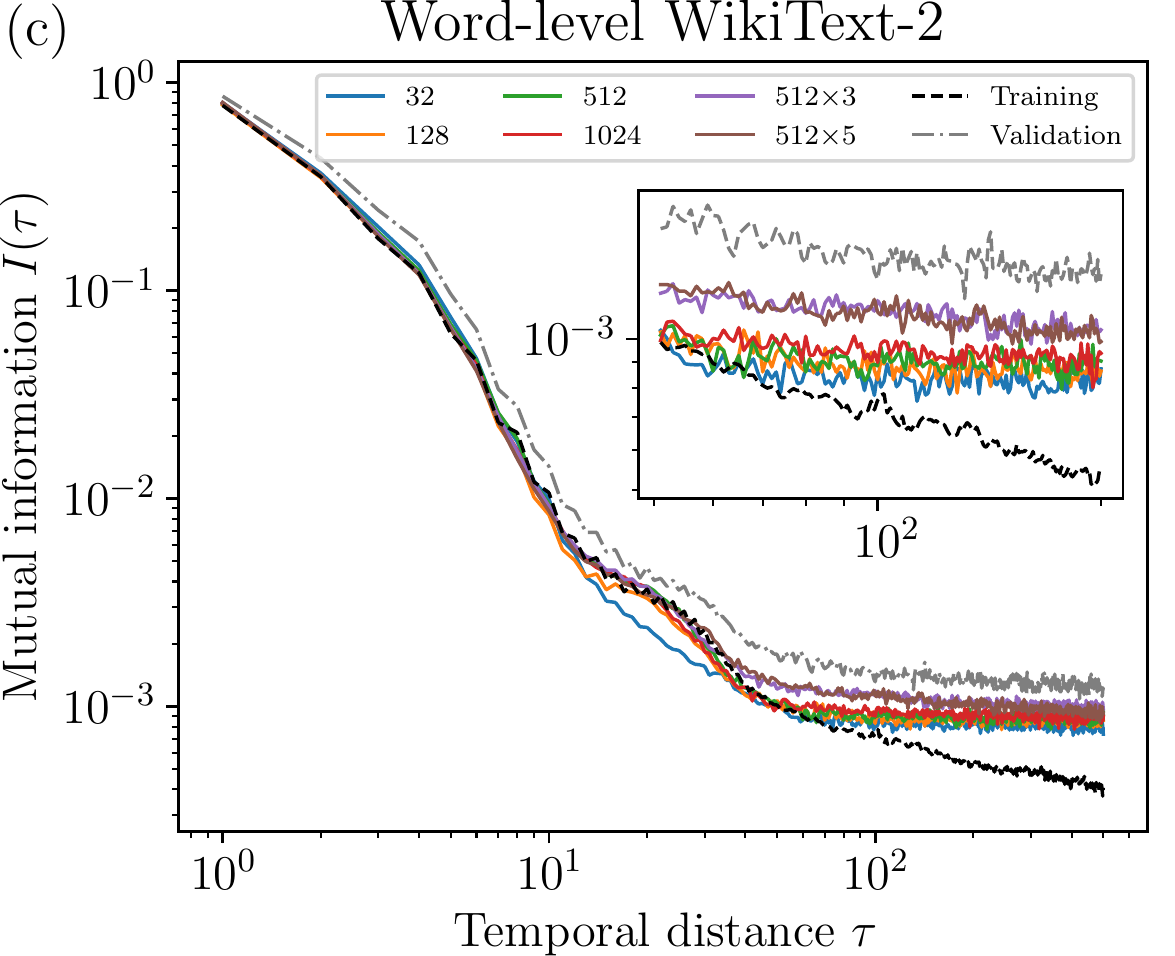}
	\hspace{4pt}
	\includegraphics[width=0.45\textwidth]{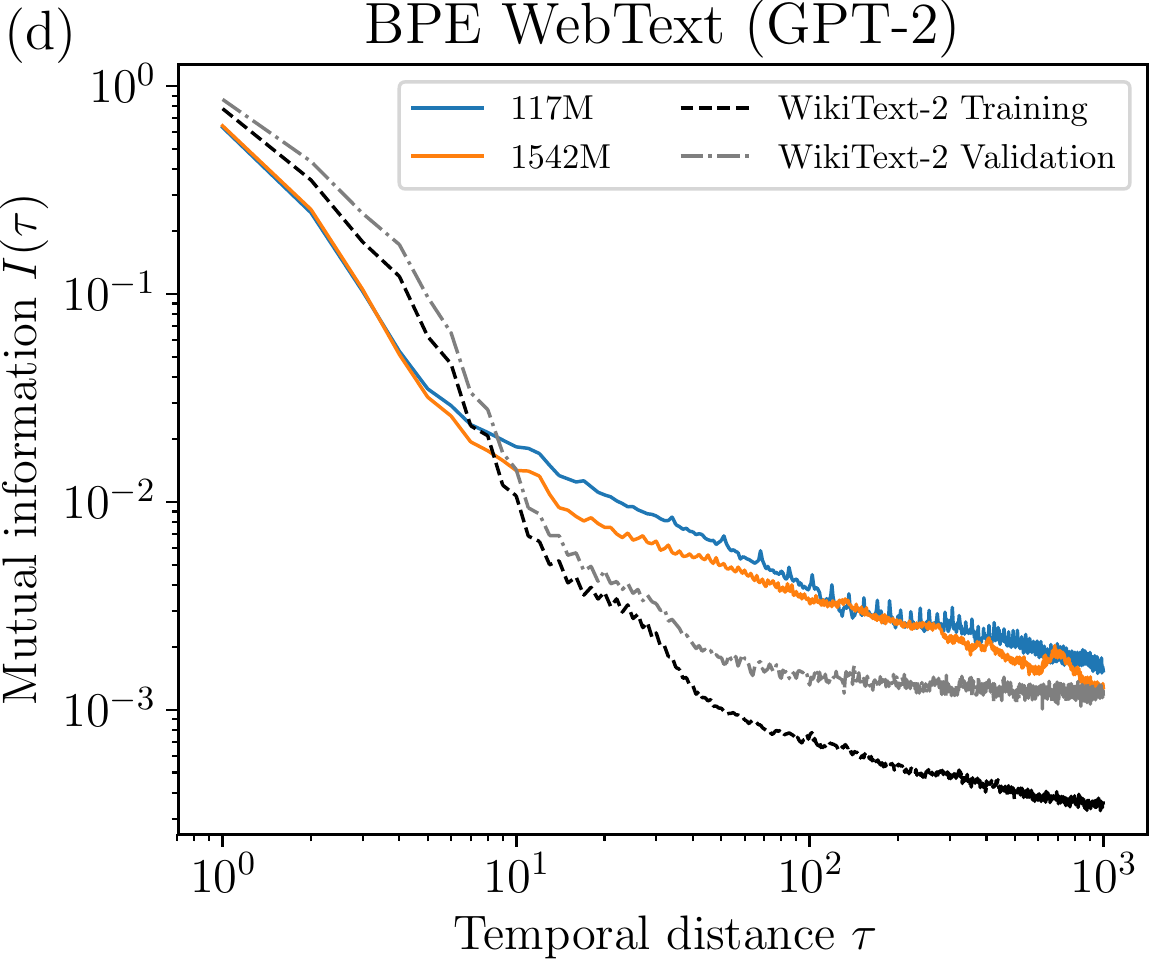}
	
	\caption{Estimated mutual information of unconditionally generated sequences from Transformers on various datasets. The legend denotes width($\times$depth, if depth$>$1). (a) Binary sequences; (b) WikiText-2 at character level; (c) WikiText-2 at word level; (d) GPT-2 model trained on WebText with byte pair encoding (BPE) \cite{sennrich-etal-2016-neural}. }
	\label{fig:trans}
\end{figure}

\subsection{Natural Language}
The mutual information of character-level Transformers already look very similar to that of word-level LSTMs. Both short and intermediate mutual information is captured and there is an overestimated power-law tail in the long distance for the single layer $m=512$ model. 
Even small word-level Transformers can track the mutual information very closely up to intermediate distances. Moreover, our three and five-layer Transformer models have a slowly decaying power-law tail up to $\tau\sim 500$, although the power does not track the training data very well. As a bonus, we evaluate the current state-of-the-art language model GPT-2 \cite{Radford2019}, where the long-range algebraic decay is persistent in both small (117M parameters) and large (1542M parameters) models. Note that the magnitude of the mutual information is much larger than that in WikiText-2, probably due to the better quality of the WebText dataset. This is beneficial to training, because it is more difficult for the network to distinguish the dependence from noise when the magnitude of the mutual information is small. Also, the discrepancy between the training and validation set of WikiText-2 may also make learning long-range dependence harder. Therefore, we speculate the bad dataset quality is the main reason why our word-level Transformer model cannot learn the power-law behavior very well on WikiText-2. 

Finally, we observe the connection between the quality of the generated text and the mutual information scaling behavior. Short-range mutual information is connected to the ability of spelling words correctly, and the intermediate-range mutual information is related to the ability to close ``braces'' in section titles correctly, or to consecutively generate two or three coherent sentences. The long-range mutual information is reflected by the long paragraphs sharing a single topic. Some text samples are presented in Appendix.

\section{Discussion}

RNNs encode the sequence history into a fixed-length and continuous-valued vector. In linear RNNs, after we ``integrate out'' the hidden state, the history dependence in Equation \eqref{eq:his} takes a form that is exponential in time. In this way, the network always ``forgets'' the past exponentially fast, thus cannot capture the power-law dependence in the sequence. In nonlinear RNNs, although we cannot analytically ``integrate out'' the hidden state, the experimental results suggest that RNNs still forget the history exponentially fast. 

It is very tempting to connect the above result to statistical mechanics. At thermal equilibrium, systems with power-law decaying mutual information or correlation are called ``critical''. It is well-known as van Hove's theorem that in one dimension, systems with short-range interactions cannot be critical, and the mutual information always decays exponentially \cite{VANHOVE1950137,ruelle1999statistical,landau2013statistical,Cuesta2004}. Importantly, here ``short range'' means the interaction strength has a finite range or decays exponentially in distance. On the other hand, one-dimensional systems with long-range interactions can be critical. A classical example is the long-range Ising model, where the finite-temperature critical point exists when the interaction strength decays as $d^{-\alpha}$, $1<\alpha<2$ \cite{dyson1969}. At exactly the critical point, the mutual information decays algebraically, and in a very vast parameter space near the critical point, the mutual information decays very slowly so that no parameter fine-tuning is needed \cite{PhysRevE.91.052703}. 

Linear RNNs resemble statistical mechanical systems with exponential decaying interactions. Therefore they cannot accommodate long-range correlations. Transformers exploit the entire sequence history explicitly. In fact, there is no natural notion of ``distance'' in Transformers, and the distance is artificially encoded using positional encoding layers. Because Transformers resemble statistical mechanical systems with long-range interactions, there is no surprise that they can capture long-range correlations well. 

An advantage of the statistical mechanical argument is its robustness. Essentially, the result claims the \emph{universality class} that a sequence model belongs to, which usually does not depend on many microscopic details. Analyzing nonlinear models and find their universality classes will be left as future work. 
However, one should be cautious that because the sampling is only conditioned on the past but not the future, the mapping between the Boltzmann distribution $p(x_1,\ldots,x_t)=e^{-\beta H(x_1,\ldots,x_t)}$ and the conditional distribution $p(x_i|x_1,\ldots,x_{i-1})$ is straightforward only in limited cases. See Appendix for a simple example. 

The implication of the above result is that self-attention networks are more efficient in capturing long-range temporal dependence. However, this ability comes at the cost of extra computational power. In order to generate a sequence of length $L$, RNNs only need  $O(L)$ time while Transformers need $O(L^2)$ (suppose no truncation on the window size). How to maintain long-range interactions while on the meantime reduce the computational complexity will be a challenging and interesting problem even from the perspective of statistical physics. It is also interesting to see how augmenting RNNs with memory can improve the long-range scaling behavior \cite{Graves2014,NIPS2015_5648,NIPS2015_5846,NIPS2015_5857,Grave2017}. 

Last but not least, the theoretical study of linear RNNs only focuses on the intrinsic expressive power of its architecture. In reality, the practical expressive power is also heavily affect by how well the stochastic gradient descent algorithm performs on the model. The fact that Transformers are long-range interacted also helps backpropagation algorithm \cite{chapter-gradient-flow-2001}. It will be interesting to connect the statistical physics to the gradient flow dynamics.

\section{Conclusion}
This paper demonstrates that RNNs are not efficient in capturing long-range temporal correlations both theoretically and empirically. Self-attention models like Transformers can capture long-range correlations much more efficiently than RNNs do, and reproduce the power-law mutual information in natural language texts. 

We also notice the non-uniformity problem in popular natural language datasets. We believe a high-quality dataset is essential for the network to learn long-range dependence.  

We hope this work provides a new perspective in understanding the expressive power of sequence models and shed new light on improving both the architecture and the training dynamics of them.

\subsubsection*{Acknowledgments}
HS thanks Yaodong Li, Tianxiao Shen, Jiawei Yan, Pengfei Zhang, Wei Hu and Yi-Zhuang You for valuable discussions and suggestions.

\bibliographystyle{ieeetr}
\bibliography{manuscript}

\newpage

\appendix
\section{Exponential Mutual Information in Linear RNNs with Gaussian Output}
In this section, we present the full proof of exponential mutual information in linear Elman RNNs with Gaussian output. For the sake of self-containedness, some of the arguments in the main text will be repeated. The proof can be roughly divided into four steps:
\begin{enumerate}
	\item Derive the recurrence relations of the block covariance matrix;
	\item Solve the recurrence relation using generating functions;
	\item Perform asymptotic analysis by studying the singularities of generating functions;
	\item Compute the mutual information based on the asymptotic analysis. 
\end{enumerate}

\paragraph{Problem Setup}
The classical Elman RNN with the linear activation is given by:
\begin{align}
h_t=&W_h x_{t-1} + U_h h_{t-1}, \label{eqapp:ht}\\
o_t=&U_o h_{t-1}. \label{eqapp:ot}
\end{align}
$o_t\in\mathbb{R}^d$ parameterizes the probability distribution from which $x_t$ is sampled, and  $h_t\in\mathbb{R}^m$ is the hidden state. In this way, the shapes of the weight matrices are $W_h\in\mathbb{R}^{m\times d}$, $U_h\in\mathbb{R}^{m\times m}$, and $U_o\in\mathbb{R}^{d\times m}$. 
In the following, we assume $x_t$ is sampled from a multivariate Gaussian distribution with mean $o_tx$ and covariance matrix proportional to the identity, i.e. $p(x_t|x_{t-1},\ldots,x_0,h_0)\sim \N(o_t,\sigma^2 I_d)$. It follows from iteratively applying Equation \eqref{eqapp:ht} that $h_t=U_h^t h_0+\sum_{i=0}^{t-1} U_h^{t-1-i} W_h x_i$. Therefore 
\begin{equation}
o_t=U_oU_h^{t}h_0+\sum_{i=0}^{t-2} U_o U_h^{t-1-i} W_h x_i. \label{eqapp:his}
\end{equation}

The joint probability distribution of the entire sequence factorizes as
\begin{align}
&p(x_t,x_{t-1},\ldots,x_0,h_0)=\prod_{i=1}^{t}p(x_i|x_{i-1},\ldots,x_0,h_0)p(h_0,x_0) \notag\\
=&\frac{1}{\sqrt{(2\pi)^{(t+2)d}\sigma^{2td}\det (\Sigma_0)}}\exp\left[-\frac{1}{2}\left((h_0-\bar{h}_0, x_0-\bar{x}_0)\Sigma_0^{-1}\begin{pmatrix}
h_0 -\bar{h}_0\\ x_0-\bar{x}_0
\end{pmatrix}+\frac{1}{\sigma^2}\sum_{i=1}^t \lVert x_i-o_i\rVert^2\right)\right], \label{eqapp:jpdf}
\end{align}
where 
\begin{equation}
\Sigma_0=\begin{pmatrix}
\Sigma_{h_0h_0} & \Sigma_{h_0x_0} \\ \Sigma_{h_0x_0}^T & \Sigma_{x_0x_0}
\end{pmatrix}.
\end{equation}
Here we have assumed $p(h_0,x_0)$ is also a multivariate Gaussian random variable. 

The hidden state is often initialized as $h_0=0$. Under this initial condition, $p(x_0)$ is multivariate Gaussian, and so does the joint distribution of the entire sequence  $p(x_t,\ldots,x_0,h_0)$. In this way, the random process is a Gaussian process. Since we are interested in the mutual information between $x_{t_0}$ and $x_{t_0+t}$ for some generic $t_0$, without loss of generality we can set $t_0=0$ and let $p(h_0,x_0)$ be a generic multivariate Gaussian distribution. We can also let $p(h_0,x_0)$ be the distribution that is already averaged over the entire sequence. In any case, we will see the asymptotic behavior of the mutual information is almost independent of $p(h_0,x_0)$. 

\paragraph{Deriving Recurrence Relations}
We are particularly interested in the distribution $p(x_0,x_t)$. Because marginalization of multivariate Gaussian random variable is still a multivariate Gaussian one, we only need to compute the block covariance matrix $\Sigma_{x_0x_t}$ and $\Sigma_{x_tx_t}$. They can be derived easily by noting $x_t=o_t+y_t$, where $y_t\sim \N(0,\sigma^2 I_d)$ and is independent. In this way,
\begin{align}
	\Sigma_{x_tx_0}=&\bE\left[x_tx_0^T\right]-\bE[x_t]\bE[x_0]^T \notag\\
	=&\bE\left[o_tx_0^T\right]-\bE[o_t]\bE[x_0]^T \notag\\
	=&U_oU_h^{t}\bE[h_0x_0^T]+\sum_{i=0}^{t-2} U_o U_h^{t-1-i} W_h \bE[x_ix_0^T]-\bE[o_t]\bE[x_0]^T \notag\\
	=&U_oU_h^{t}\left(\Sigma_{h_0x_0}+\bE[h_0]\bE[x_0]^T\right)+\sum_{i=0}^{t-2} U_o U_h^{t-1-i} W_h \left(\Sigma_{x_ix_0}+\bE[x_i]\bE[x_0]^T\right)-\bE[o_t]\bE[x_0]^T \notag\\
	=&U_oU_h^{t}\Sigma_{h_0x_0} +\sum_{i=0}^{t-2}U_oU_h^{t-1-i}W_h\Sigma_{x_ix_0}, \label{eqapp:rec1}
\end{align}
where in the second line we have used the decomposition $x_t=o_t+y_t$ and the independence of $y_t$, and in the last line we have used the fact that
\begin{equation}
	\bE[o_t]=U_oU_h^{t}\bE[h_0]+\sum_{i=0}^{t-2} U_o U_h^{t-1-i} W_h \bE[x_i].
\end{equation}
Similarly, 
\begin{align}
	\Sigma_{x_tx_t}=&\bE[x_tx_t^T]-\bE[x_t]\bE[x_t]^T \notag\\
	=&\bE[y_ty_t^T]+\bE[o_to_t^T]-\bE[o_t]\bE[o_t]^T \notag\\
	=&\sigma^2I_d+\Sigma_{o_t o_t}. \label{eqapp:rec2}
\end{align}

As a sanity check, Equation \eqref{eqapp:rec1} and \eqref{eqapp:rec2} can also be derived directly from the probability density function Equation \eqref{eqapp:jpdf}. $\Sigma_{x_t x_t}$ and $\Sigma_{x_0x_t}$ only depend on $\Sigma_{x_t'x_t'}$ and $\Sigma_{x_0x_t'}$ for $t'<t$. First rewrite Equation \eqref{eqapp:jpdf} as the canonical form of multivariate Gaussian distribution:
\begin{align}
p(x_t,x_{t-1},\ldots,x_0,h_0)=&\frac{1}{\sqrt{(2\pi)^{(t+2)d}\det(\Sigma)}}\exp\left[-\frac{1}{2}(\mathbf{x}-\bar{x}_0)^T\Sigma^{-1}(\mathbf{x}-\bar{x}_0)\right],
\end{align}
where $\mathbf{x}=(h_0,x_0,x_1,\ldots,x_t)^T$ and $\Sigma$ is a symmetric matrix. The last row of $\Sigma^{-1}$ is
\begin{equation}
\frac{1}{\sigma^2}\begin{pmatrix}
-U_oU_h^{t}  & -U_oU_h^{t-1}W_h & -U_oU_h^{t-2}W_h  & \ldots & -U_oU_h W_h & 0 & I_d
\end{pmatrix},
\end{equation}
and the second column of $\Sigma$ is
\begin{equation}
\begin{pmatrix}
\Sigma_{h_0x_0} \\ \Sigma_{x_0x_0} \\ \Sigma_{x_1x_0} \\  \vdots \\ \Sigma_{x_{t-1}x_0} \\ \Sigma_{x_tx_0}
\end{pmatrix}.
\end{equation}
Since $\Sigma^{-1}\Sigma =I_{(t+2)d}$, the product of the last row  of $\Sigma^{-1}$ and the second column of $\Sigma$ should be zero. This gives 
\begin{equation}
\Sigma_{x_tx_0}=U_oU_h^{t}\Sigma_{h_0x_0} +\sum_{i=0}^{t-2}U_oU_h^{t-1-i}W_h\Sigma_{x_ix_0}. 
\end{equation}
Similarly, by considering the last row of $\Sigma^{-1}$ and the last column of $\Sigma$, we obtain
\begin{equation}
\Sigma_{x_tx_t}=\sigma^2I_d+U_oU_h^{t}\Sigma_{h_0x_t}+\sum_{i=0}^{t-2}U_oU_h^{t-1-i}W_h\Sigma_{x_ix_t}.
\end{equation}
%
%It is also helpful to consider the last row of $\Sigma^{-1}$ and the first column of $\Sigma$, which gives 
%\begin{equation}
%\sigma^2\Sigma_{x_th_0}=U_oU_h^{t}\Sigma_{h_0h_0} +\sum_{i=0}^{t-2}U_oU_h^{t-1-i}W_h\Sigma_{x_ih_0}, \label{eqapp:rec3}
%\end{equation}
%which is almost identical to Equation \eqref{eqapp:rec1} except that in the non-homogeneous term $\Sigma_{h_0x_0}$ is replaced by $\Sigma_{h_0h_0}$. 

\paragraph{Solving Recurrence Relations}
We solve the linear recurrence relation of $\Sigma_{x_t x_0}$, given by Equation \eqref{eqapp:rec1}. 
Define the auxiliary sequence 
\begin{equation}
A_n=\begin{cases}
0, & n=1, \\
U_oU_h^{n-1} W_h, & n\geq 2.
\end{cases}
\end{equation}
and the following three formal power series:
\begin{align}
\Sigma(z)\equiv&\sum_{n=0}^{\infty}\Sigma_{x_nx_0} z^n, \\
A(z)\equiv&\sum_{n=1}^{\infty} A_n z^n, \\
B(z)\equiv&\sum_{n=1}^{\infty} U_oU_h^{n}\Sigma_{h_0x_0} z^n.
\end{align}
With the identity
\begin{equation}
\left(\sum_{m=0}^\infty B_m z^m\right)\left(\sum_{n=0}^\infty A_n z^n\right)=\sum_{n=0}^\infty\left(\sum_{m=0}^n B_{n-m}A_m\right) z^n,
\end{equation}
Equation \eqref{eqapp:rec1} is equivalent to
\begin{equation}
\Sigma(z)-\Sigma_{x_0x_0}=B(z)+A(z)\Sigma(z),
\end{equation}
or
\begin{equation}
\Sigma(z)=\left[I_d-A(z)\right]^{-1}\left[B(z)+\Sigma_{x_0x_0}\right]. \label{eqapp:s}
\end{equation}

To this end, we assume the square matrix $U_h$ bears an eigenvalue decomposition $U_h=Q \Lambda Q^{-1}$, where $\Lambda=\mathrm{diag}(\lambda_1,\ldots,\lambda_m)$  with $|\lambda_1|\geq\ldots\geq |\lambda_m|$ and $Q$ is an orthogonal matrix. With this assumption, we can obtain the closed form expression of $A(x)$ and $B(x)$:
\begin{align}
A_{ij}(z)=&\sum_{n=2}^{\infty}\sum_{k=1}^m (U_oQ)_{ik}\lambda_k^{n-1}(Q^{-1}W_h)_{kj} z^n \notag\\
=&\sum_{k=1}^m (U_oQ)_{ik}\left(\sum_{n=2}^{\infty} \lambda_k^{n-1}z^n \right) (Q^{-1}W_h)_{kj} \notag\\
=&\sum_{k=1}^m   (U_oQ)_{ik} \frac{\lambda_k z^2}{1-\lambda_k z }(Q^{-1}W_h)_{kj},
\end{align}
or in matrix form
\begin{align}
A(z)=&U_oQ(I_m-\Lambda z)^{-1}\Lambda Q^{-1}W_h z^2 \notag\\
=&U_o(I_m-U_h z)^{-1}U_hW_h z^2. 
\label{eqapp:ax}
\end{align}
Similarly, 
\begin{align}
B_{ij}(z)=&\sum_{n=1}^{\infty}\sum_{k=1}^m (U_oQ)_{ik}\lambda_k^{n}(Q^{-1}\Sigma_{h_0x_0})_{kj} z^n \notag\\
=&\sum_{k=1}^m (U_oQ)_{ik}\left(\sum_{n=1}^{\infty}\lambda_k^{n} z^n\right)(Q^{-1}\Sigma_{h_0x_0})_{kj} \notag\\
=&\sum_{k=1}^m (U_oQ)_{ik}\frac{\lambda_k z}{1-\lambda_k z}(Q^{-1}\Sigma_{h_0x_0})_{kj},
\end{align}
or in matrix form
\begin{align}
B(z)=&U_oQ(I_m-\Lambda z)^{-1} \Lambda Q^{-1}\Sigma_{h_0x_0} z \\
=&U_o(I_m-U_h z)^{-1} U_h\Sigma_{h_0x_0} z. \label{eqapp:bx}
\end{align}
Insert Equation \eqref{eqapp:ax} and \eqref{eqapp:bx} into \eqref{eqapp:s}, we obtain the formal solution for $\Sigma(z)$:
\begin{align}
\Sigma(z)=\left[I_d-U_o(I_m-U_h z)^{-1}U_h W_h z^2\right]^{-1}\left[U_o(I_m-U_h z)^{-1} U_h\Sigma_{h_0x_0} z+\Sigma_{x_0x_0}\right]. \label{eqapp:sigx}
\end{align}

%Note that because the recurrence relation of  $\Sigma_{x_th_0}$ is almost identical to that of $\Sigma_{x_tx_0}$, the formal solution to 
%\begin{equation}
%	\Sigma'(z)=\sum_{n=0}^{\infty}\Sigma_{x_0h_n} z^n, 
%\end{equation}
%is
%\begin{equation}
%\Sigma'(z)=\left[I_d-\frac{1}{\sigma^2}U_oQ(I_m-\Lambda z)^{-1}\Lambda Q^{-1}W_h z^2\right]^{-1}\left[\frac{1}{\sigma^2}U_oQ(I_m-\Lambda z)^{-1} \Lambda Q^{-1}\Sigma_{h_0h_0} z+\Sigma_{x_0h_0}\right]. \label{eqapp:sigp}
%\end{equation}

\paragraph{Asymptotic Analysis}
The long time asymptotic behavior of $\Sigma_{x_0x_t}$ can be analyzed by treating $\Sigma(z)$ as a function on the complex plane and studying it singularities \cite{flajolet2009analytic}. Since matrix inversion can be computed from Cramer's rule, $\Sigma(z)$ is a rational function of $z$, and its singularities always occur as poles. Therefore, the asymptotic behavior of $\Sigma_{x_tx_0}$ will be exponential in $t$, whose exponent is determined by the position of the pole closet to the origin. The order of the pole determines the polynomial sub-exponential factor. 

Because we are dealing with a matrix here, each element has its own set of poles. Denote the pole closest to the origin associated with $\Sigma_{ij}(z)$ as $z^0_{ij}$ and its order as $n_{ij}\in\mathbb{N}^+$. In this way, $(\Sigma_{x_tx_0})_{ij}\sim t^{n_{ij}-1}|z^0_{ij}|^{-t}$ when $t$ is large. Unless $z^0_{ij}$ is exactly on the unit circle, which is of zero measure in the parameter space, the solution either increases or decreases exponentially with $t$. Even if $z^0_{ij}$ is on the unit circle, the power of the polynomial can only be an integer. If any element in $\Sigma_{x_tx_0}$ increases exponentially with $t$, $\Sigma_{x_tx_0}$ simply remembers the initial condition, which is not desirable. Therefore, in any working network every element of $\Sigma_{x_tx_0}$ decreases exponentially in $t$. %The same result also applies for $\Sigma_{h_tx_0}$ and we denote the pole closest to the origin associated with $\Sigma'_{ij}(z)$ as $z'^0_{ij}$. 

The pole position $z^0_{ij}$ is bounded by only $W_h$, $U_h$ and $U_o$, independent of the initial condition $p(x_0,h_0)$. For example, consider the term proportional to $\Sigma_{x_0x_0}$ in Equation \eqref{eqapp:sigx}:
\begin{equation}
\Sigma_{ij}(z)=\sum_{k=1}^d F_{ik}(z)\left(\Sigma_{x_0x_0}\right)_{kj}, \label{eqapp:f}
\end{equation}
where $F(z)\equiv\left[I_d-U_o(I_m-U_h z)^{-1} U_hW_h z^2\right]^{-1}$. Equation \eqref{eqapp:f} is exactly the partial fraction decomposition of a rational function. (Of course, $F_{ik}(z)$ may be decomposed further.)
In this way, $z^0_{ij}$ is the pole closest to the origin, among all poles of $F_{ik}(z)$ and for all $k$, unless $\left(\Sigma_{x_0h_0}\right)_{kj}$ for that particular $k$ is exactly zero, in which case the resulting pole will be further away from the origin, leading to a faster decay. 

We then analyze the asymptotic behavior of $\Sigma_{x_tx_t}$. The only property that we need about it is non-degeneracy. Intuitively, this is true because it is sampled conditionally from a Gaussian distribution with covariance matrix $\sigma^2I_d$. Formally, according to Courant minimax principle, the minimal eigenvalue of $\Sigma_{x_tx_t}$ is given by
\begin{equation}
	\min_{\lVert x\rVert=1}x^T \left(\Sigma_{x_tx_t}\right)x=\sigma^2+\min_{\lVert x\rVert=1} x^T \left(\Sigma_{o_to_t}\right) x\geq \sigma^2,
\end{equation}
where we have inserted Equation \eqref{eqapp:rec2} and used the fact that $\Sigma_{o_to_t}$ is positive semi-definite. In this way, $\lim_{t\to\infty}\Sigma_{x_tx_t}=O\Gamma O^T $, where $\Gamma=\mathrm{diag}(\gamma_1,\ldots,\gamma_d)$, $\gamma_1>\ldots>\gamma_d\geq \sigma^2>0$ is the eigenvalue of the covariance matrix and $O$ is an orthogonal matrix.

\paragraph{Mutual Information Computation}
Let $X\in\mathbb{R}^n$ , $Y\in\mathbb{R}^m$ be two multivariate Gaussian random variables with a joint distribution $p_{XY}\sim\N_{n+m}(\mu,\Sigma)$, where
\begin{equation}
\mu=\begin{pmatrix}
\mu_X \\ \mu_Y
\end{pmatrix},\ \Sigma=\begin{pmatrix}
\Sigma_{XX} & \Sigma_{XY} \\ \Sigma_{XY}^T & \Sigma_{YY}
\end{pmatrix}.
\end{equation} 
Use one of the definitions of the mutual information, $I(X;Y)=H(X)+H(Y)-H(X,Y)$, the entropy of multivariate Gaussian random variables $H(X)=\frac{1}{2}\ln \det(2\pi e \Sigma)$ and the Schur decomposition
\begin{equation}
\begin{split}
\det(\Sigma)=&\det(\Sigma_{YY})\det(\Sigma_{XX}-\Sigma_{XY}\Sigma_{YY}^{-1}\Sigma_{XY}^T)\\
=&\det(\Sigma_{XX})\det(\Sigma_{YY}-\Sigma_{XY}^T\Sigma_{XX}^{-1}\Sigma_{XY}).
\end{split}
\end{equation}
we have
\begin{equation}
\begin{split}
I(X;Y)=&\frac{1}{2}\ln \frac{\det (\Sigma_{YY})}{\det(\Sigma_{YY}-\Sigma_{XY}^T\Sigma_{XX}^{-1}\Sigma_{XY})}=-\frac{1}{2}\ln  \det(I_m-\Sigma_{YY}^{-1}\Sigma_{XY}^T\Sigma_{XX}^{-1}\Sigma_{XY})\\
=&\frac{1}{2}\ln \frac{\det (\Sigma_{XX})}{\det(\Sigma_{XX}-\Sigma_{XY}\Sigma_{YY}^{-1}\Sigma_{XY}^T)}=-\frac{1}{2}\ln  \det(I_n-\Sigma_{XX}^{-1}\Sigma_{XY}\Sigma_{YY}^{-1}\Sigma_{XY}^T).
\end{split}
\end{equation}

Using the above formula, the mutual information between $x_0$ and $x_t$ can then be computed as
\begin{equation}
I(x_0;x_t)
=-\frac{1}{2}\ln \det(I_d-\Sigma_{x_0x_0}^{-1}\Sigma_{x_0x_t}\Sigma_{x_tx_t}^{-1}\Sigma_{x_0x_t}^T)
\approx\frac{1}{2}\mathrm{tr}(\Sigma_{x_0x_0}^{-1}\Sigma_{x_0x_t}\Sigma_{x_tx_t}^{-1}\Sigma_{x_0x_t}^T). \label{eqapp:tr}
\end{equation}
$\Sigma_{x_0x_0}$ is time-independent. Each element in $\Sigma_{x_0x_t}$ decays exponentially with $t$. Because the minimal eigenvalue of $\Sigma_{x_tx_t}$ is bounded by $\sigma^2$, $\Sigma_{x_tx_t}^{-1}$ is well-defined and tends to a finite constant matrix $\lim_{t\to\infty}\Sigma_{x_tx_t}^{-1}=O\Gamma^{-1}O^T$. In this way, elements in $\Sigma_{x_0x_0}^{-1}\Sigma_{x_0x_t}\Sigma_{x_tx_t}^{-1}\Sigma_{x_0x_t}^T$ is exponentially small in the large $t$ limit, which justifies the last equality in Equation \eqref{eqapp:tr}. Because trace is a linear function, the mutual information decreases exponentially with $t$. This finishes the proof that in any linear Elman RNN with Gaussian output that does not simply memorize the initial condition, the mutual information decays exponentially with time. The only technical assumption we made in the proof is that the covariance matrix $\Sigma_{x_tx_0}$ decreases exponentially in time instead of exponentially, without which the network simply memories the initial condition. 

We note that adding bias terms in Equation \eqref{eq:ht} and \eqref{eq:ot} does not affect the conclusion because the mutual information of Gaussian random variables only depends on the covariance matrix, while the bias terms only affect their mean. 

%Finally, we briefly discuss the magnitude of the decay exponent. The pole position $z_{ij}^0$ is in general closer to the origin than $1/|\lambda_1|$, where $\lambda_1$ the largest eigenvalue in magnitude in $\Lambda$. In other words, $\Sigma_{x_tx_0}$ decays slower than $|\lambda_1|^{t}$. Notice $F(z)\equiv\left[I_d-U_o(I_m-U_h z)^{-1}U_h W_h z^2\right]^{-1}$ factor in Equation \eqref{eqapp:sigx}. 
%Assume none of the poles in $(I_m-U_h z)^{-1}$ vanishes due to the additional factors $U_o$ and $U_h W_h $. In this way, $(F^{-1})_{ij}(0)=\delta_{ij}$ and $(F^{-1})_{ij}(\lambda_1)=-\infty$. By continuity, there must exist some $z_0^{ij}$ between the origin and $|\lambda_1|$, such that $F^{-1}(z_0^{ij})=0$, which is a pole of $\Sigma_{ij}(z)$. In the literature of analytic combinatorics, this type of sequence is called ``supercritical'' \cite{flajolet2009analytic}. 

\section{Binary Sequence Generation from Multivariate Gaussian}
In this section, we report a method to sample fixed-length sequences of binary random variables with arbitrary designated mutual information. The method is used to generate the binary sequence dataset used in this paper. 

Consider a bivariate normal random variable $X\sim\N_2(0,\Sigma)$ with mean zero and covariance matrix
\begin{equation}
\Sigma\equiv \begin{pmatrix}
\Sigma_{11} & \Sigma_{12} \\
\Sigma_{21} & \Sigma_{22}
\end{pmatrix}\equiv \begin{pmatrix}
a & c \\
c & b
\end{pmatrix},
\end{equation}
where $a,b>0$ due to the non-negativity of the variance. The condition that the covariance matrix is positive semi-definite is that $\det(\Sigma)= ab-c^2\geq 0$. 

Define the bivariate Bernoulli random variable $Y\equiv\mathrm{sgn}(X)$, where the sign function applies element-wisely. The mutual information between $Y_1$ and $Y_2$ is
\begin{equation}
I(Y_1;Y_2)=\sum_{\alpha,\beta=\pm 1}p_{Y_1Y_2}(\alpha,\beta)\ln\frac{p_{Y_1Y_2}(\alpha,\beta)}{p_{Y_1}(\alpha)p_{Y_2}(\beta)}=\ln 4+\sum_{\alpha,\beta=\pm 1}p_{Y_1Y_2}(\alpha,\beta)\ln p_{Y_1Y_2}(\alpha,\beta),
\end{equation}
where we have used the fact that $p_{Y_1}(\alpha)=p_{Y_2}(\beta)=1/2$ for all $\alpha,\beta=\pm 1$. Although an analytical  expression for general $a,b,c$ is viable, for simplicity, we take $a=b=1$ and $c=\cos\theta>0$ such that $\theta\in [0,\pi/2]$. Straightforward integration yields
\begin{align}
p_{Y_1Y_2}(0,0)=p_{Y_1Y_2}(1,1)=&\frac{1}{2}-\frac{\theta}{2\pi} \\
p_{Y_1Y_2}(1,0)=p_{Y_1Y_2}(0,1)=&\frac{\theta}{2\pi},
\end{align}
The mutual information $I(Y_1;Y_2)$ as a function of $\theta$ is then 
\begin{equation}
I(\theta)=\frac{\pi\ln(2/\pi)+(\pi-\theta)\ln(\pi-\theta)+\theta\ln\theta}{\pi}. \label{eqapp:it}
\end{equation}
As a sanity check, when $c=0$ or $\theta=\pi/2$, due to the property of multivariate normal random variable, $X_1$ and $X_2$ become independent and so do $Y_1$ and $Y_2$. This is consistent with the fact that $I(\pi/2)=0$. When $c=1$ or $\theta=0$, $X_1$ and $X_2$ become perfectly correlated such that $X_1=X_2$ and thus $Y_1=Y_2$. This is consistent with the fact that $I(0)=\ln 2$. 

Now consider a multivariate normal random variable $X\sim\N_N(0,\Sigma)$. Because of the nice marginalization property of $X$, namely, for any subspace of the random vector $X$: $X_p=(X_{i_1},\ldots,X_{i_p})$, the marginal distribution of $X_p$ is still a multivariate normal random variable, with mean zero and covariance matrix $\Sigma_{ij\in\{i_1,\ldots,i_p\}}$. Therefore, we can independently control the covariance between any pair of $X_i$ and $X_j$ and thus the mutual information between any pair of $Y_i$ and $Y_j$ with $Y\equiv\mathrm{sgn}(X)$. 

For example, in order to engineer a sequence with mutual information decaying asymptotically as power law: $I(Y_{i};Y_{i+d})=Ad^{-\gamma}$ for all $i=0,\ldots,N-d-1$, we simply let 
\begin{equation}
\Sigma_{ij}=\frac{\sqrt{A/2}\pi}{|i-j|^{\gamma/2}}, \label{eqapp:sig}
\end{equation}
for all $i\neq j$. 
Expanding $I(c)$ around $c=0$ we have $I(c)=2c^2/\pi^2+o(c^2)$. Therefore when $d$ is large the asymptotic mutual information of the sequence is $I(d)\approx Ad^{-\gamma}$. If the mutual information is required to obey power law even when $d$ is small, one should invert Equation \eqref{eqapp:it} as a function of $\theta$. Since we are mostly interested in large $d$ scaling behavior of the mutual information, this inversion is not necessary. 

\section{Constant Mutual Information in Repetitive Sequences}
In this section, we show how short-range memory effect can cause non-decaying mutual information. 

\begin{figure}[H]
	\centering
	\includegraphics[width=0.45\textwidth]{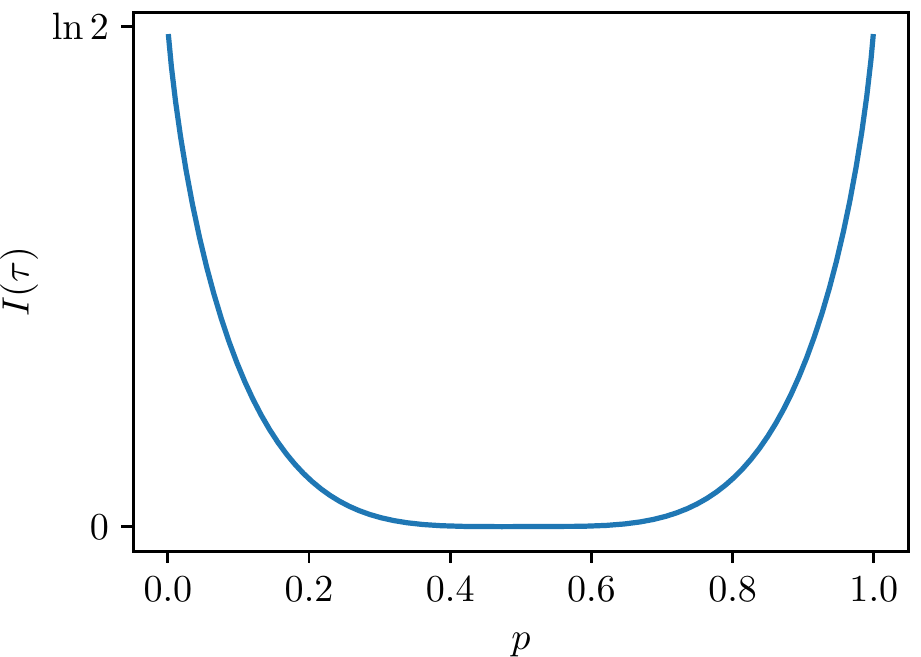}
	\caption{Plot of the auto-mutual information in Equation \eqref{eqapp:i}. }
	\label{figapp:short}
\end{figure}

Consider an infinite long sequence $\{x_t:t\in\mathbb{N}\}$, where the $2n$-th and $(2n+1)$-th symbol are 01 with probability $p$ and 10 with probability $1-p$ for all $n\in \mathbb{N}$. We would like to compute the auto-mutual information in this sequence. For each symbol, 
\begin{equation}
p(0)=p(0|\text{even } n)p(\text{even } n)+p(0|\text{odd } n)p(\text{odd } n)=p\cdot\frac{1}{2}+(1-p)\cdot\frac{1}{2}=\frac{1}{2},
\end{equation}
and thus $p(1)=1-p(0)=1/2$. We would like to compute $p(x_t,x_{t+\tau})$ for $\tau>1$. Depending on whether $\tau$ is even or odd, there are two scenarios. Here we first assume $\tau$ is even. In this case:
\begin{align}
&p(x_t=0,x_{t+\tau}=0) \notag\\
=&p(x_t=0,x_{t+\tau} =0|\text{even }t)p(\text{even }t)+p(x_t=0,x_{t+\tau}=0|\text{odd }t)p(\text{odd }t) \notag\\
=&\frac{1}{2}\left[p^2+(1-p)^2\right].
\end{align}
Similarly, 
\begin{align}
&p(x_t=0,x_{t+\tau}=0)=p(x_t=1,x_{t+\tau}=1)=\frac{1}{2}\left[p^2+(1-p)^2\right], \\
&p(x_t=0,x_{t+\tau}=1)=p(x_t=1,x_{t+\tau}=0)=p(1-p). 
\end{align}
In this way, the auto-mutual information is
\begin{align}
I(\tau)=&\sum_{i,j=0}^1 p(x_t=i,x_{t+\tau}=j)\ln\frac{p(x_t=i,x_{t+\tau}=j)}{p(x_t=i)p(x_{t+\tau}=j)} \notag\\
=&4p(p-1)\mathrm{arctanh}[(1-2p)^2]+\ln[2+4p(p-1)].  \label{eqapp:i}
\end{align}
It is not hard to see the scenario when $\tau>1$ is odd is completely symmetric. The functional form is plotted in Figure \ref{figapp:short}, which is nonzero as long as $p\neq 1/2$. In this way, the mutual information is non-decaying constant. 

Therefore, as long as there exists some bias in generating different short symbol patterns, the long-range mutual information of the sequence does not decay to complete zero. 

From the statistics point of view, this sequence can be generated from a three-state HMM: $p(A\to B)=p$, emit symbol 0; $p(A\to C)=1-p$, emit symbol 1; $p(B\to A)=1$, emit symbol 1; $p(C\to A)=1$, emit symbol 0. Note that this HMM is periodic and the mutual information does not need to decay exponentially.  

From the statistical physics point of view, this resembles systems in ferromagnetic or anti-ferromagnetic ordered phase, where short-range interactions lead to magnetic domain and non-decaying long-range correlation. However, due to the short-range nature of the interaction, one cannot infer more useful information from the mutual information other than the domain. 

\section{Mutual Information Estimation}
\subsection{Mutual Information Estimation}
Unbiased estimation of entropy is a nontrivial problem. Here we use the entropy estimator proposed in Ref.~\cite{Grassberger2008}:
\begin{equation}
\hat{H}=\ln N-\frac{1}{N}\sum_{i=1}^V n_i\psi(n_i),
\end{equation}
where $n_i$, $i=1,\ldots,V$ is the number of observations of the $i$-th symbol, $V$ is the number of unique symbols, and $\psi(z)$ is the digamma function. $N=\sum_{i=1}^V n_i$. The mutual information is then estimated as
\begin{equation}
\hat{I}(X;Y)=\hat{H}(X)+\hat{H}(Y)-\hat{H}(X,Y).
\end{equation}

\begin{figure}[tbp]
	\centering
	\includegraphics[width=0.45\textwidth]{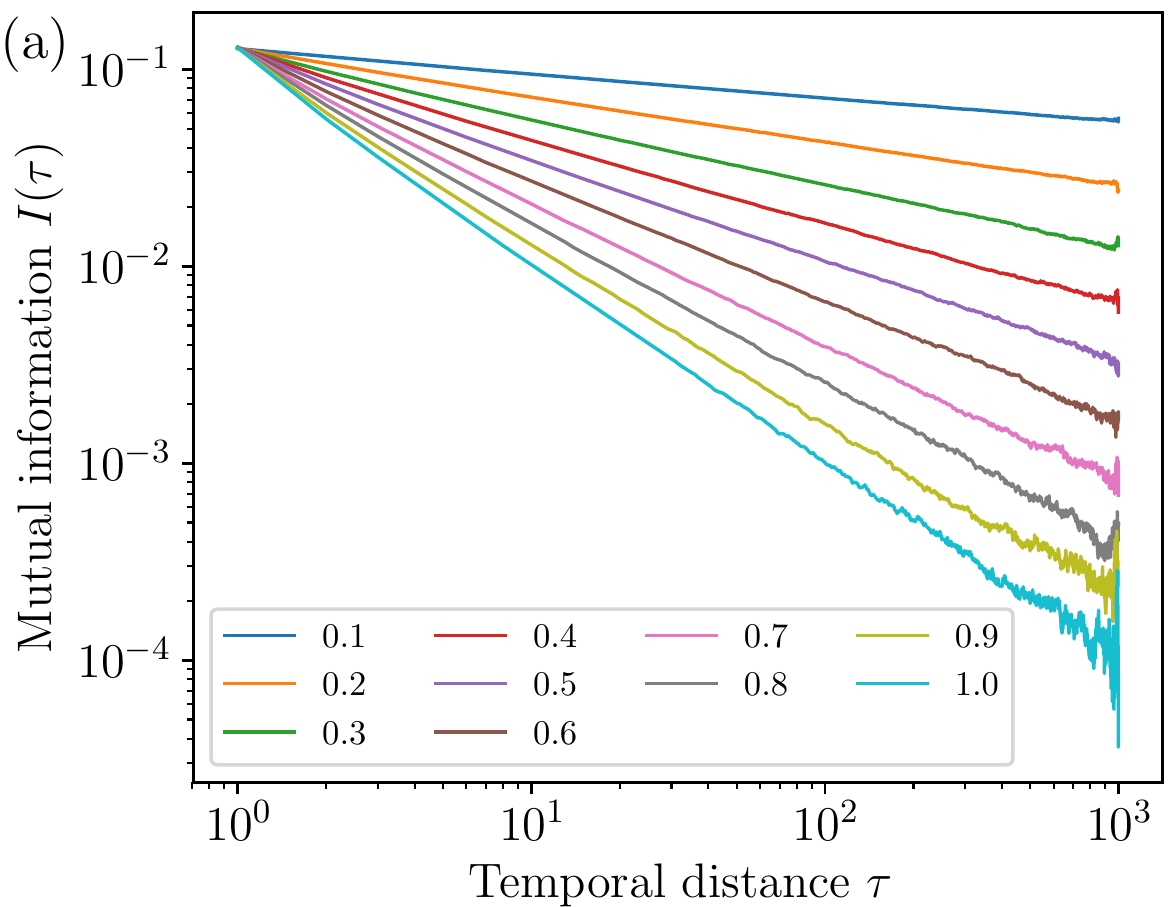}
	\hspace{4pt}
	\includegraphics[width=0.45\textwidth]{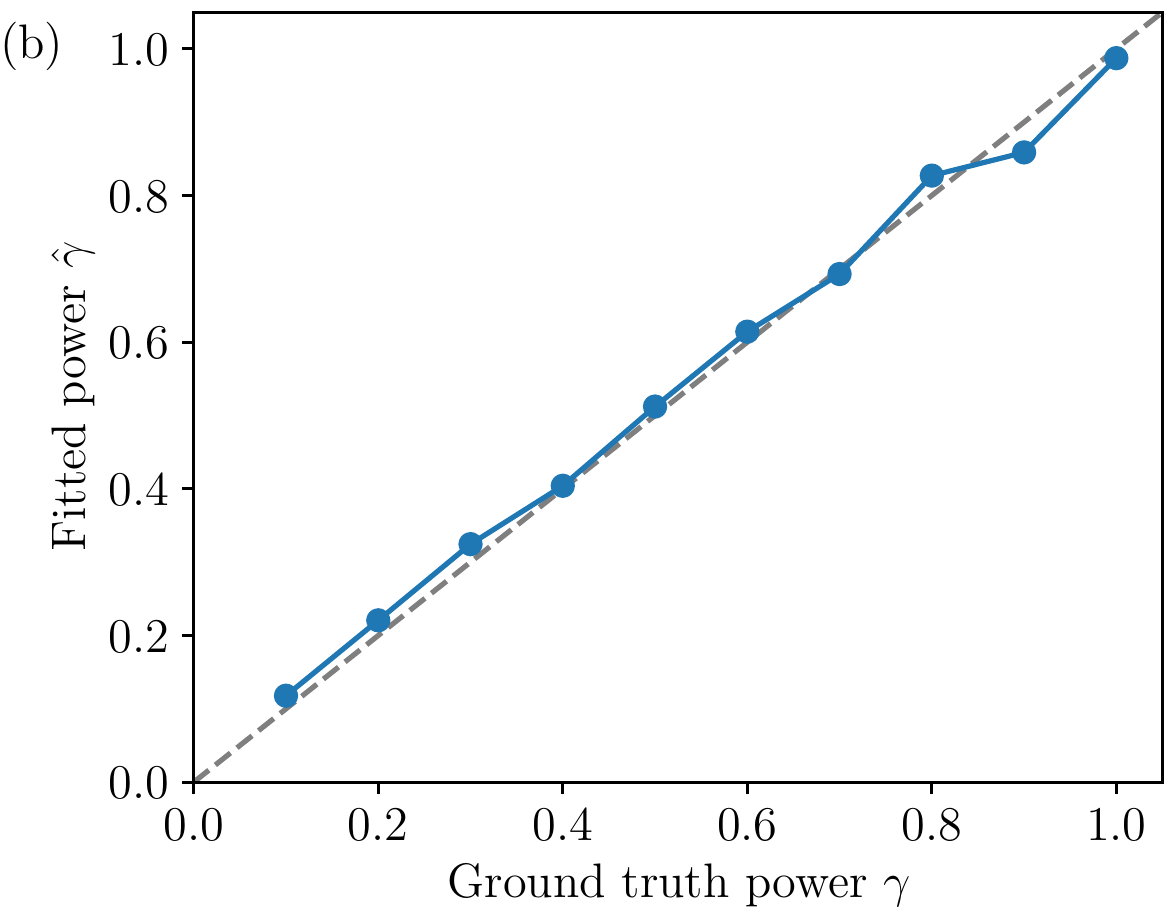}
	\caption{(a) Estimated mutual information of binary sequences with mutual information $I(\tau)=0.1\tau^{-\gamma}$ and different power $\gamma$; (b) Fitted power $\hat{\gamma}$ versus ground truth $\gamma$. The error bar that represents the 95\% confidence interval is smaller than the size of the blue marker. }
	\label{figapp:bench}
\end{figure}

We benchmark this estimator on the artificial binary sequences introduced in this paper. We first generate 10000 sequences of length 1000, whose auto-mutual information decays as $I(\tau)=0.1\tau^{-\gamma}$, where $\gamma$ ranges from 0.1 to 1. We then estimate the auto-mutual information using the above estimator and fit the power $\gamma$. The results are shown in Figure \ref{figapp:bench}. The bottom right part of Figure \ref{figapp:bench}(a) is a little bit noisy because (i) there are less samples used for estimating mutual information in the long distance; (ii) the logarithmic scale enhances relative statistical errors when the mutual information is small. As can be seen, the estimator can estimate the mutual information quite accurately. 

\subsection{Mutual Information on Natural Language Datasets}
We estimate the mutual information \textit{at character level} on Penn Treebank \cite{marcus-etal-1993-building}, WikiText-2, WikiText-103 \cite{Merity2016} and text8 \cite{text8}. For text8, we use the consecutive 90MB, 5MB, 5MB of text as the training, validation and test set respectively. 

Although the power-law decaying mutual information is observed in all the training sets, it is missing in the validation and  test sets of Penn Treebank, WikiText-2 and WikiText-103. The text8 dataset, where the data is split by ourselves, is more uniform.

\begin{figure}[tbp]
	\centering
	\includegraphics[width=0.45\textwidth]{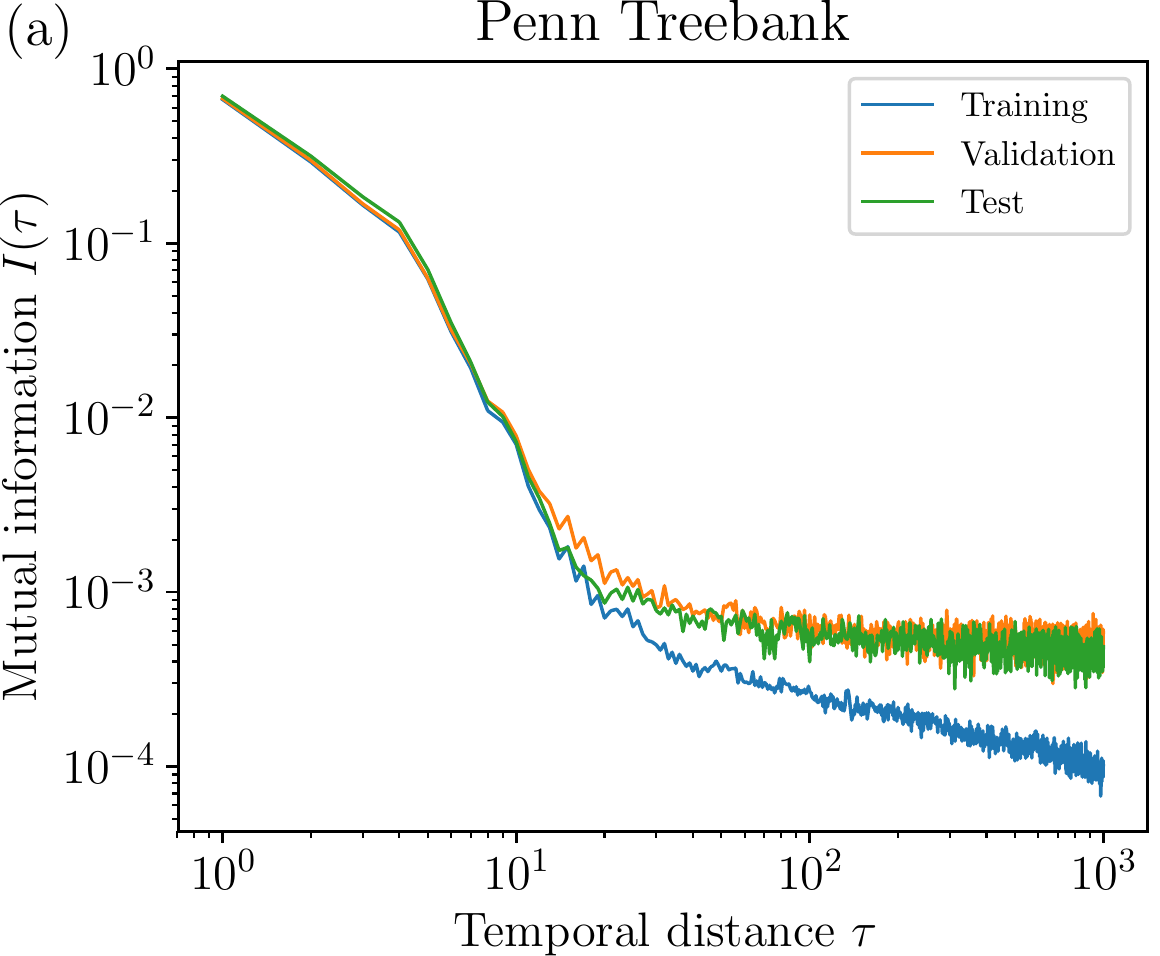}
	\hspace{4pt}
	\includegraphics[width=0.45\textwidth]{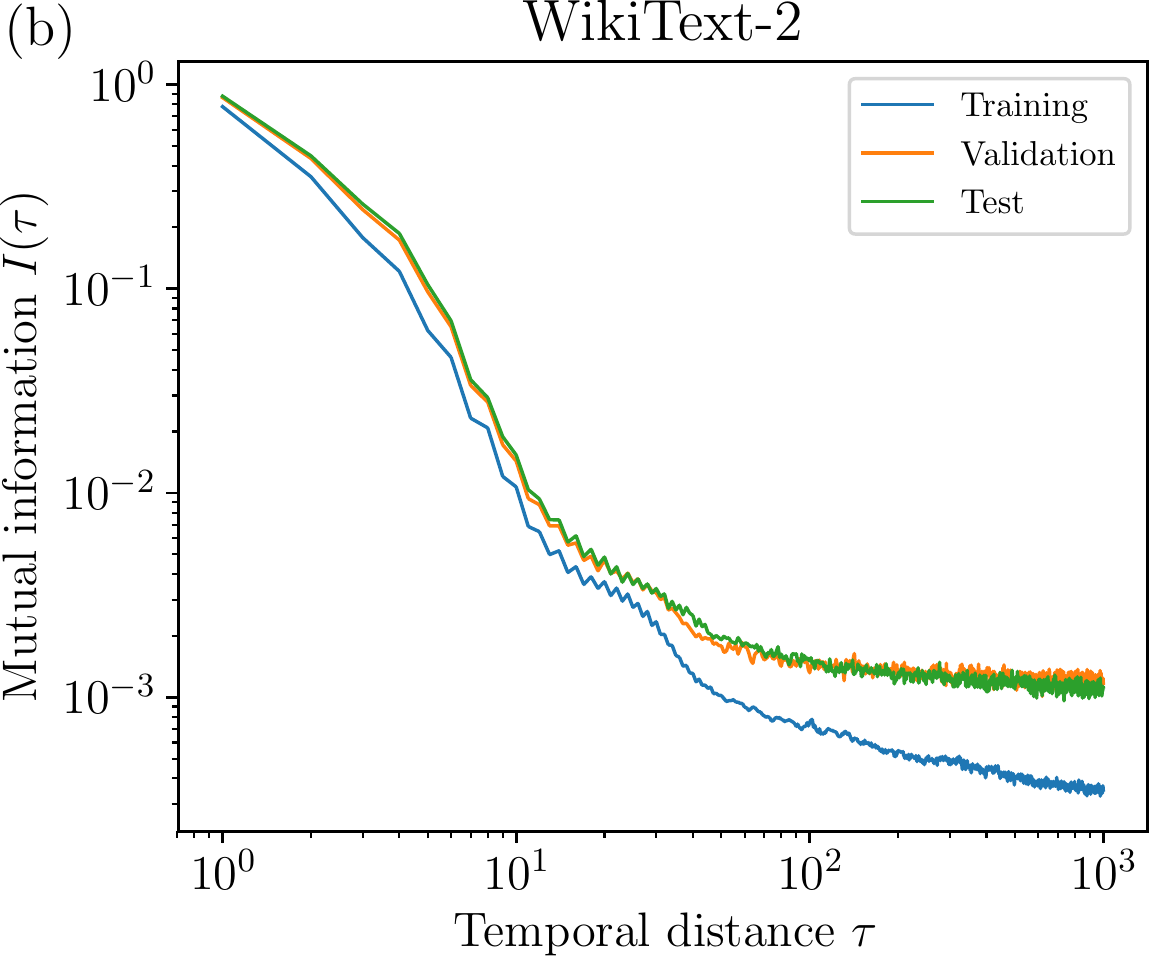}
	
	\vspace{8pt}
	\includegraphics[width=0.45\textwidth]{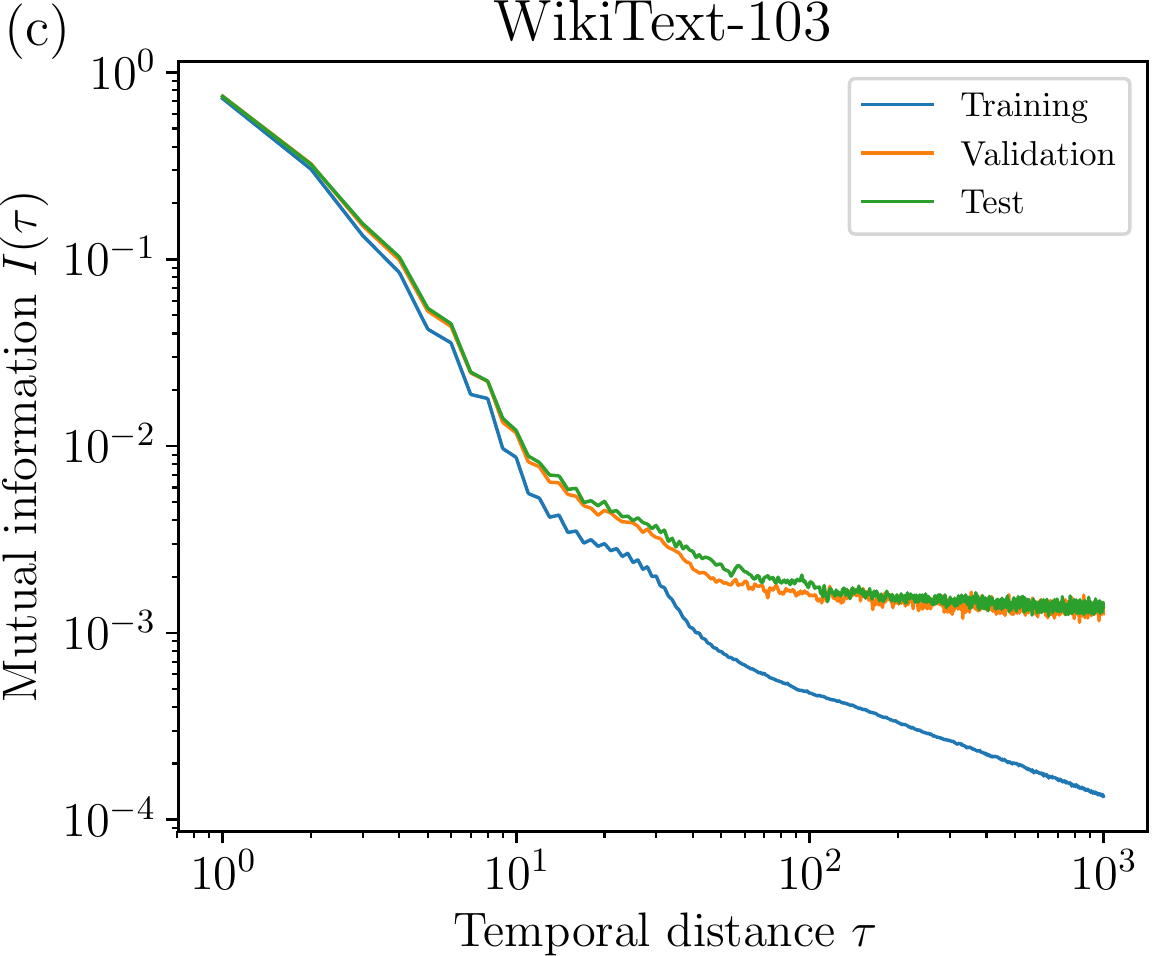}
	\hspace{4pt}
	\includegraphics[width=0.45\textwidth]{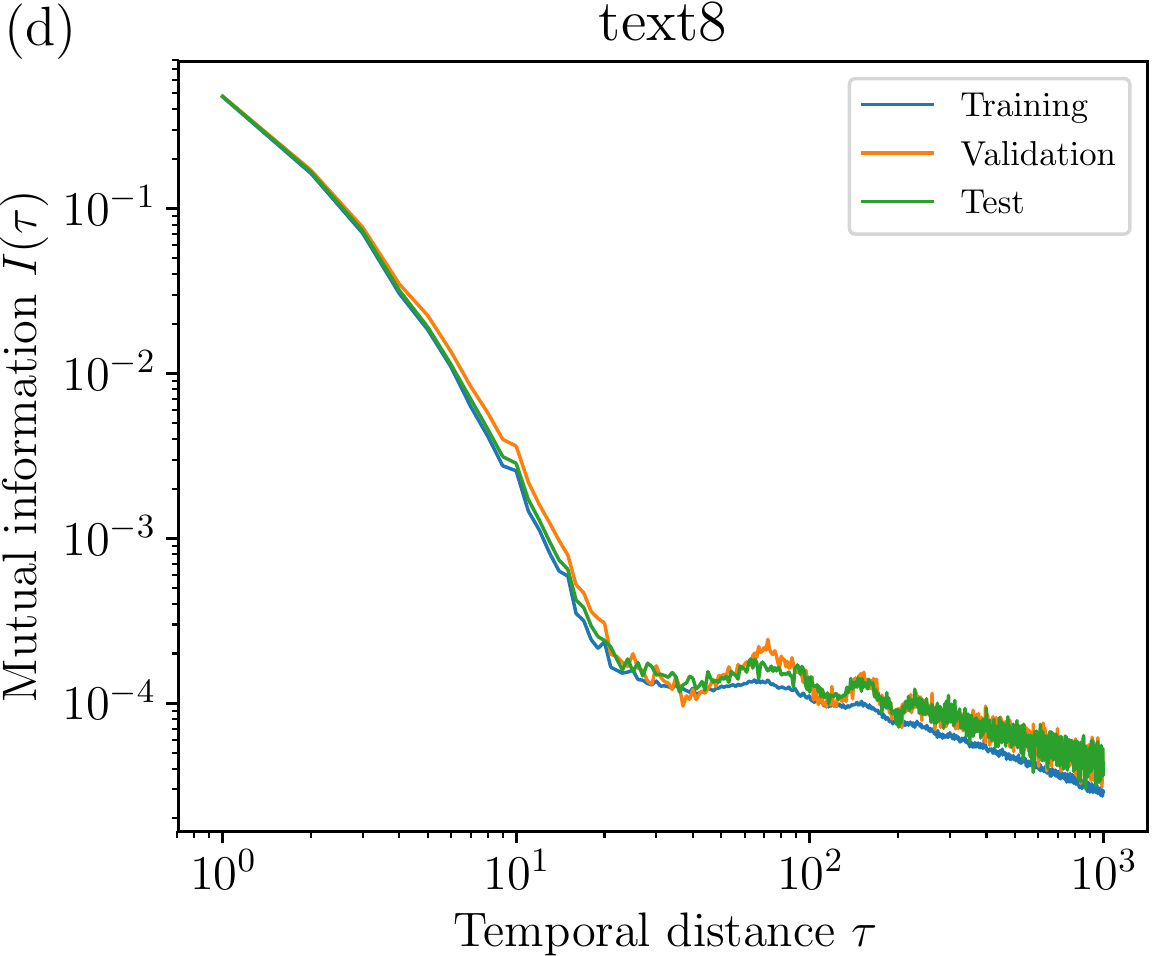}
	\caption{Estimated mutual information of various popular natural language datasets. }
	\label{figapp:datasets}
\end{figure}

\section{Experiment Details and More Results}
\subsection{RNNs}
\subsubsection{Binary Sequence}
In all RNNs, the first layer after the input layer is an embedding layer of size 2. The embedding layer is followed by RNN layers. Connected to the last RNN layer, the last fully-connected layer is of size 2 with softmax activation. 

The batch size is 50 and the training set is shuffled during the training. We do not truncate BPTT on this dataset. We use RMSProp optimizer \cite{rmsprop} and the gradient is clipped to 0.25. For Vanilla RNNs, the learning rate is 0.001. For LSTMs, the learning rate is 0.005. 

All the networks are trained long enough until the validation loss (evaluated on 2000 similar sequences) is not improving in the recent 5 epochs. To minimize the randomness introduced by the stochastic gradient descent, we run the training 5 times and generate the sequence using the model with the minimal validation loss. 

The experiments are conducted using Keras with Tensorflow backend. All parameters without specification are default in Keras. For example, the kernel initializer of RNNs is Glorot uniform \cite{pmlr-v9-glorot10a}, the recurrent initializer is orthogonal \cite{Saxe2013} and the bias is initialized to zero. 

\begin{figure}[tbp]
	\centering
	\includegraphics[width=0.45\textwidth]{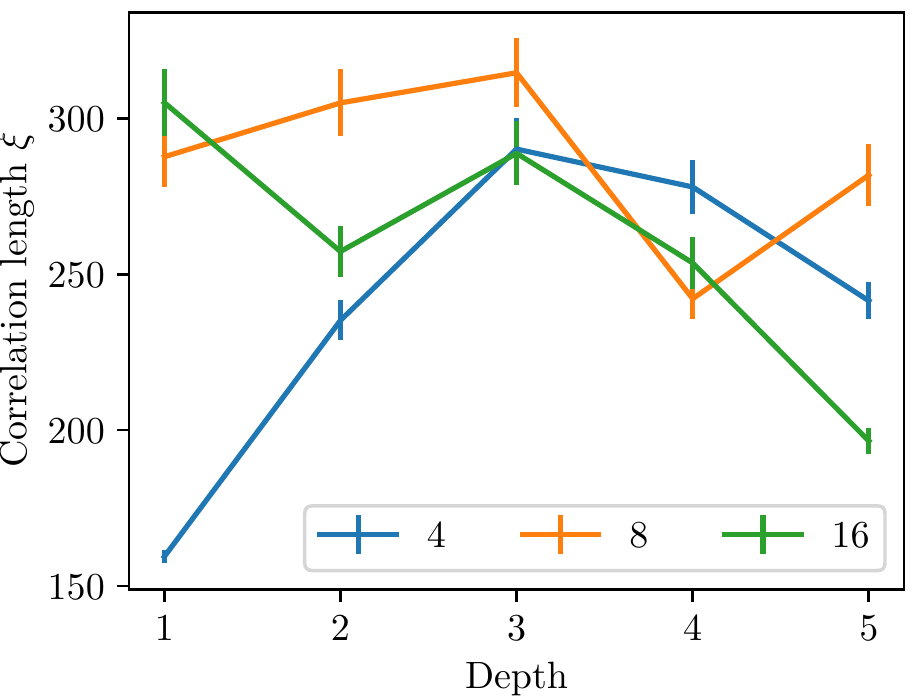}
	\caption{Fitted correlation length $\xi$ as a function of the LSTM layer depth and width on binary sequences. The legend denotes width. Only data points greater than $10^{-3}$ is used due to the estimation error at long distances. The error bar represents the 95\% confidence interval. }
	\label{figapp:binlstm}
\end{figure}

In Figure \ref{figapp:binlstm}, we supplement more data points for the depth scaling of LSTMs (only $m=8$ curve is reported in the main text due to the space limitation). The data support the fact that increasing LSTM depth over two layers does not improve the network performance. 

The better performance of LSTMs cannot be naively attributed to more trainable parameters. For example, the correlation length of  $m=32$ vanilla RNN is shorter than the $m=4$ LSTM. The former has 1190 trainable parameters while the latter has only 126.

\subsubsection{Natural Language}
\paragraph{Character level}
There are 283 unique characters in WikiText-2. In all RNNs, the first layer after the input layer is an embedding layer of size 200 with orthogonal initializer \cite{Saxe2013}. The embedding layer is followed by LSTM layers. Connected to the last LSTM layer, the last fully-connected layer is of size 283 with softmax activation. 

The batch size is 128 and the training set is \emph{not} shuffled during the training in order to keep the long-range dependence. BPTT is truncated to 100 characters. To still introduce some randomness in the training and to bias the network to be able to generate text from the zero hidden state, the hidden state of the network is reset to zero at the beginning of each batch with probability 0.01. We use RMSProp optimizer and the gradient is clipped to 0.25. The learning rate is 0.005. 

All the networks are trained long enough until the validation loss is not improving in the recent 5 epochs. To minimize the randomness introduced by the stochastic gradient descent, we run the training 5 times and generate the sequence using the model with the minimal validation loss. 

The experiments are conducted using Keras with Tensorflow backend. All parameters without specification are default values in Keras.

\paragraph{Word level}
There are around 33K unique tokens in WikiText-2. Because the vocabulary size is so large, the weight of the first embedding layer and the last fully-connected layer is tied to reduce the model size \cite{press-wolf-2017-using,Inan2016}. In this way, the embedding dimension and the hidden unit dimension of LSTMs are identical. 

Because word level LSTM language model is very common, we simply use the official example code of PyTorch\footnote{\url{https://github.com/pytorch/examples/tree/master/word_language_model}}. BPTT is truncated to 50 tokens. After the embedding layer and all LSTM layers except for the last one, there are dropout layers with dropout rate 0.4. All parameters without specification are default values. For example, the optimizer is the vanilla stochastic gradient with initial learning rate 20. The learning rate is annealed by a factor of 4 if there is no improvement on the validation loss after each epoch. The gradient is clipped to 0.25. 

The model is trained 60 epochs. To minimize the randomness introduced by the stochastic gradient descent, we run the training 3 times and generate the sequence using the model with the minimal validation loss. 

\begin{figure}[tbp]
	\centering
	\includegraphics[width=0.45\textwidth]{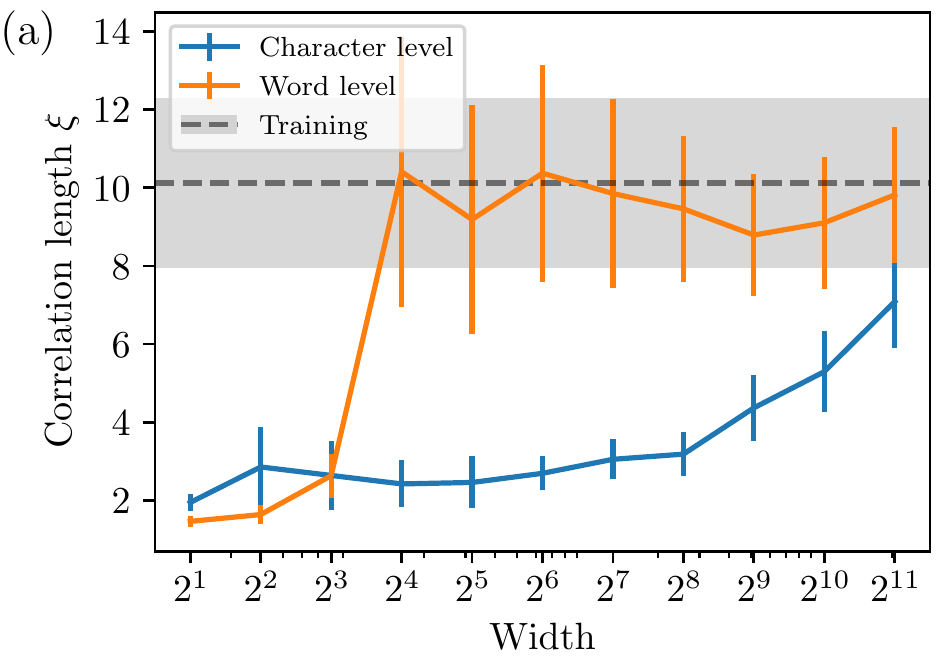}
	\hspace{4pt}
	\includegraphics[width=0.45\textwidth]{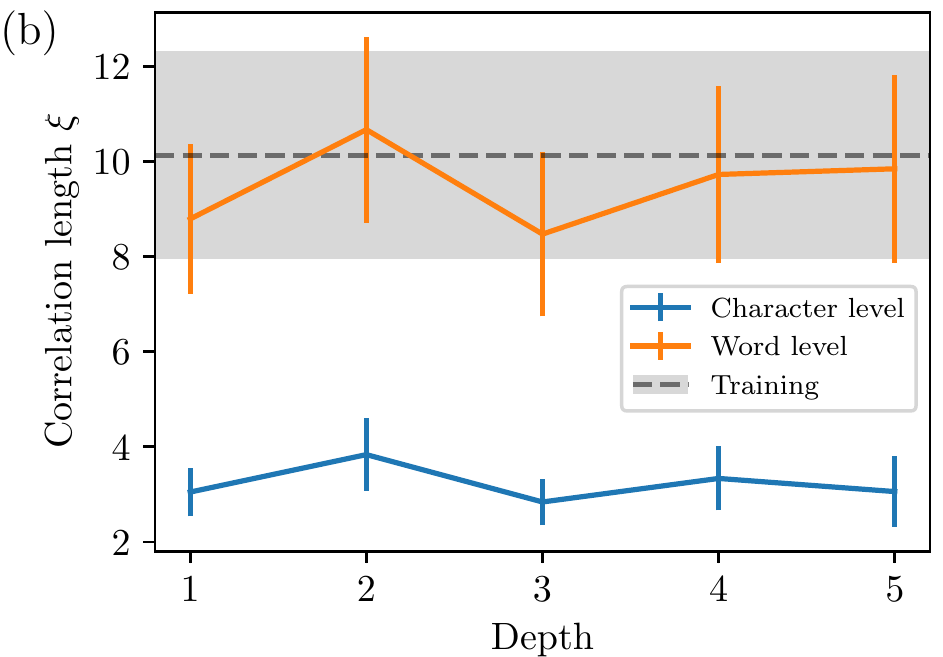}
	\caption{(a) Fitted correlation length $\xi$ as a function of the LSTM width on WikiText-2. The depth of all LSTMs is one. Only data points greater than $2\times 10^{-3}$ is used which includes features up to $\tau\lesssim 50$. The error bar represents the 95\% confidence interval. The gray dashed line represents the same fitting from the training data and the shaded area represents the 95\% confidence interval; (b) Same as (a) but as a function of the LSTM depth. The width of character-level networks is 128 and that of word-level networks is 512. }
	\label{figapp:textlstm}
\end{figure}

In Figure \ref{figapp:textlstm}, we show the scaling analysis of LSTM models by fitting the correlation length using the mutual information in short and intermediate distances. The character-level model cannot capture intermediate-range features unless it is extremely wide ($m=2048$). On the other hand, small word-level model ($m=16$) can already capture intermediate-range features quite well. The depth dependence is typical of LSTM models---increasing the depth does not improve the model significantly, if there is any improvement at all. 

\subsection{Transformers}
Because the multi-head attention layer uses residue connection, the network width 
$m=\text{number  of heads}\times\text{hidden dimension of each head}$ must equal the embedding dimension. In all Transformers, the first layer after the input layer is an embedding layer whose size is the same as the Transformer width. After the embedding layer is the additive sinusoidal and cosinusoidal positional encoding layer same as that in the original Transformer proposal \cite{NIPS2017_7181}. The positional encoding is followed by the multi-head attention layers. Connected to the last attention layer is a fully-connected layer with softmax activation. 

We use Adam optimizer \cite{Kingma2014} with $\beta_1=0.9$ and $\beta_2=0.999$, with gradient clipped to 5.0. We also apply cosine annealing to the learning rate \cite{Loshchilov2016}, where the minimal learning rate is 1/32 of the maximum learning rate. 

The Transformer model is implemented in Keras\footnote{Based on the implementation \url{https://github.com/kpot/keras-transformer}. }.

\subsubsection{Binary Sequence}
The multi-head attention layer has four heads. The batch size is 50 and the training set is shuffled during the training. The initial learning rate of Adam is 0.0002. All the networks are trained up to 200 epochs or until the validation loss is not improving in the recent 5 epochs. To minimize the randomness introduced by the stochastic gradient descent, we run the training 5 times and generate the sequence using the model with the minimal validation loss.

\begin{figure}[tbp]
	\centering
	\includegraphics[width=0.45\textwidth]{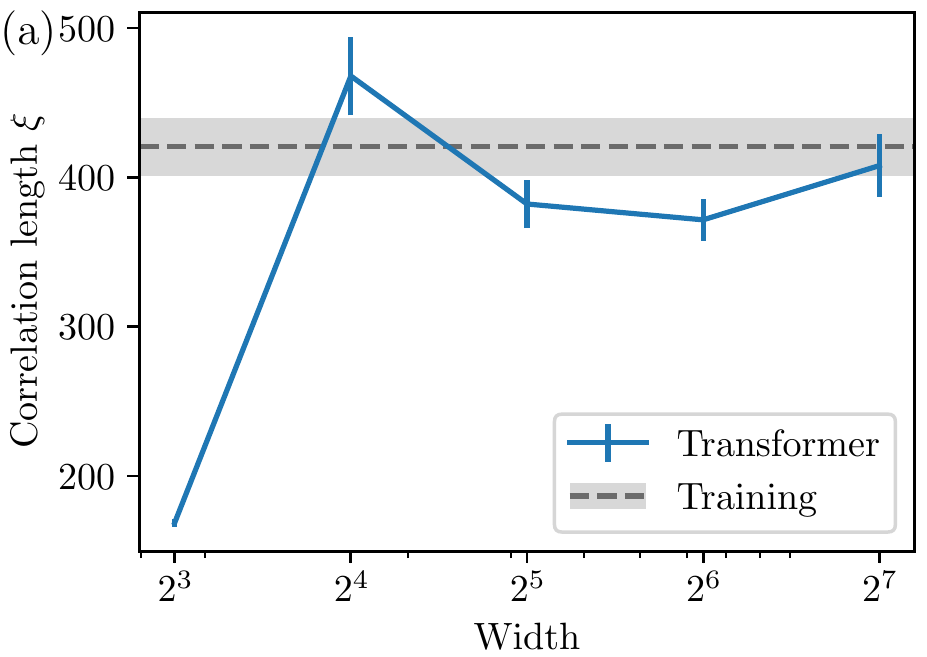}
	\hspace{4pt}
	\includegraphics[width=0.45\textwidth]{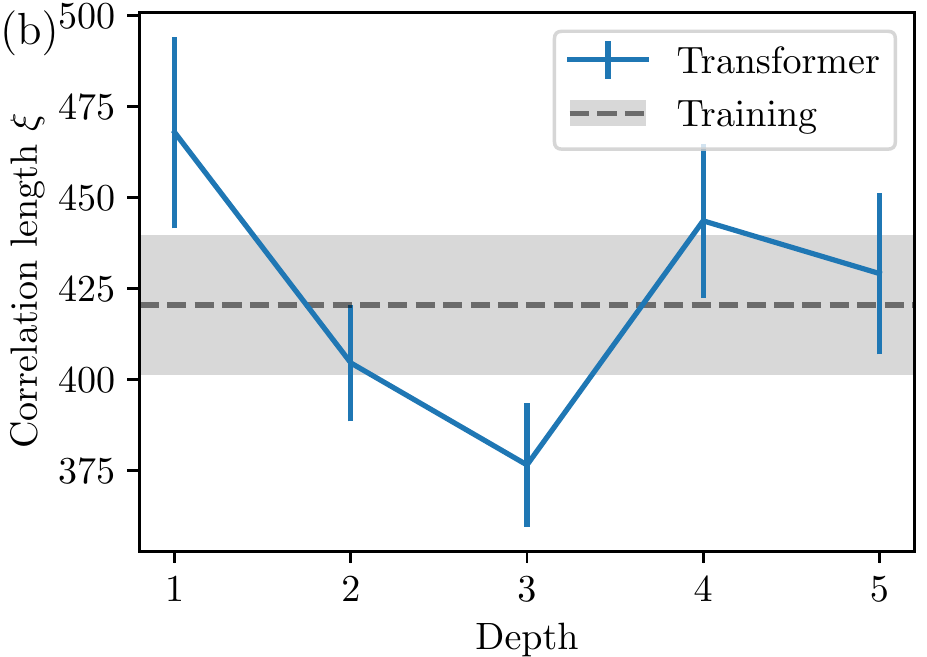}
	\caption{(a) Fitted correlation length $\xi$ as a function of the Transformer width on binary sequences. The depth of all networks is one. Only data points greater than $1\times 10^{-3}$ is used. The error bar represents the 95\% confidence interval. The gray dashed line represents the same fitting from the training data and the shaded area represents the 95\% confidence interval; (b) Same as (a) but as a function of the Transformer depth. The width of all networks is 16. }
	\label{figapp:bintrans_corr}
\end{figure}

In Figure \ref{figapp:bintrans_corr} we fit the correlation length similar to the analysis in Figure 1(c) and (d) in the main text, although the mutual information does not fit quite well with the exponential function. Note that the correlation length in our best LSTM is smaller than 350. The same fitting on the training set yields the baseline correlation length 420. The $m=16$ single layer model overestimates the mutual information and also the correlation length. 

\subsubsection{Natural Language}
\paragraph{Character level}
The multi-head attention layer has eight heads. The batch size is 64 and the training set is shuffled during the training. The initial learning rate of Adam optimizer is 0.0002. All the networks are trained up to 200 epochs or until the validation loss is not improving in the recent 5 epochs. To minimize the randomness introduced by the stochastic gradient descent, we run the training 3 times and generate the sequence using the model with the minimal validation loss.

\paragraph{Word level}
The weight of the first embedding layer and the last fully-connected layer is tied to reduce the model size \cite{press-wolf-2017-using,Inan2016}. The multi-head attention layer has eight heads. We introduce 0.1 dropout rate for all the self-attention layer, and additional 0.01 $L_2$ regularization for the embedding layer. The batch size is 20 or 32 depending on the model size. The training set is shuffled during the training. The initial learning rate of Adam optimizer is 0.00025. All the networks are trained up to 200 epochs or until the validation loss is not improving in the recent 10 epochs.

\begin{figure}[tbp]
	\centering
	\includegraphics[width=0.45\textwidth]{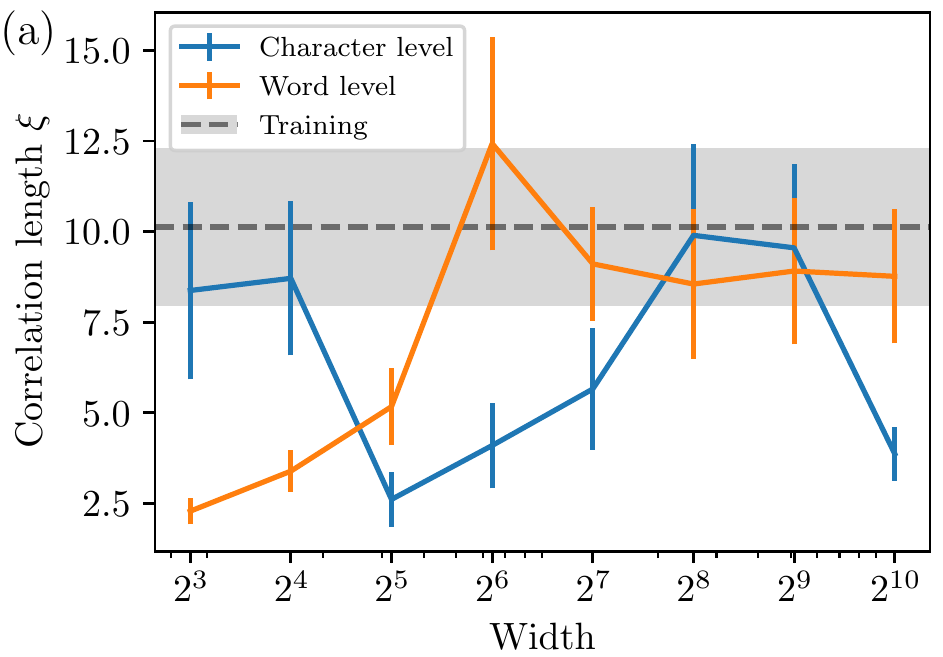}
	\hspace{4pt}
	\includegraphics[width=0.45\textwidth]{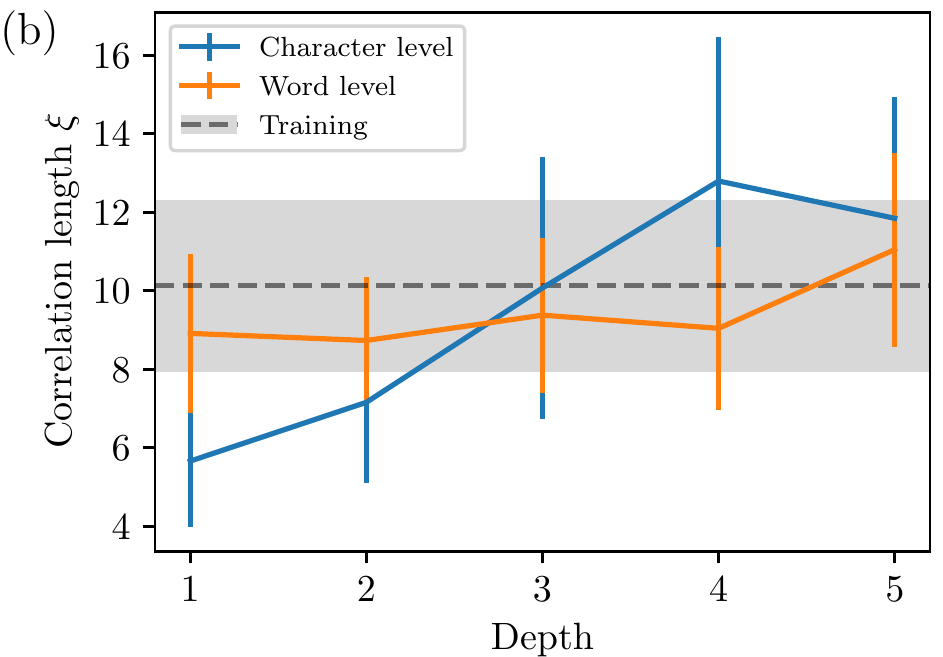}
	\caption{(a) Fitted correlation length $\xi$ as a function of the Transformer width on WikiText-2. The depth of all Transformers is one. Only data points greater than $2\times 10^{-3}$ is used which includes features up to $\tau\lesssim 50$. The error bar represents the 95\% confidence interval. The gray dashed line represents the same fitting from the training data and the shaded area represents the 95\% confidence interval; (b) Same as (a) but as a function of the Transformer depth.  The width of character-level networks is 128 and that of word-level networks is 512. }
	\label{figapp:chartrans}
\end{figure}

In Figure \ref{figapp:chartrans}, we show the scaling analysis by fitting the correlation length using the mutual information in short and intermediate distances. Different from character-level LSTMs, character-level Transformers can capture the mid-range interactions well. Also, deep Transformers perform better than shallow Transformers. 

We would like to comment on the spurious long correlation lengths on shallow Transformers in Figure \ref{figapp:chartrans}(a). As can be seen from Figure 3(b) in the main text, shallow Transformers cannot even capture the short-range mutual information well, in the sense the magnitude of $I(\tau)$ is much smaller than that of the training set. The small magnitude is not reflected by the correlation length, and does not contradict with the fact that the small magnitude can decay very slowly.

\paragraph{GPT-2}
The mutual information of GPT-2 \cite{Radford2019} is based on 500 unconditionally generated text samples of 1024 tokens. We generate samples using the pretrained 117M model. Since there is no 1542M model available to download, we simply use the provided text. We remove the sample divider
\begin{quote}
	{\tiny ======================================== SAMPLE X ========================================} 
\end{quote}
The code and the text sample can be found online\footnote{\url{https://github.com/openai/gpt-2}}. 

%\subsection{An Overfitting Analysis}
%As mentioned in the main text, because of the non-uniformity of the WikiText-2 dataset, the fact that we always pick the best model on the validation set for sequence generation may prevent the model from learning the long-range mutual information. To test whether the discrepancy indeed plays an negative role in the training, we deliberately let our networks overfit and use the overfitted network to generate sequences. 

%We study both word-level LSTM and word-level Transformer. Both models have only 

\subsection{Closing Remarks}
Although the goal of this work is to study the intrinsic expressive power of sequence model architectures, in the experiments the expressive power is inevitably limited by how well the models are trained. To remedy the randomness in the training, we typically train several models and select the best model for sequence generation. 

Limited by computational resources, we do not fine-tune the training hyperparameters. We also note there exist more sophisticated training algorithms \cite{Merity2017}. It would be definitely interesting to perform the same scaling analysis on models trained by them, although we speculate the performance will improve only quantitatively but not qualitatively. 

\section{Mutual Information and Text Quality}
In this section we present some generated text samples that show the connection between the mutual information and the text quality. 
\begin{itemize}
	\item Character-level LSTM with $m=8$, single layer. Even short-range mutual information is not completely captured. 
	\begin{quote}
		\small 
		\textit{Por hsaantre Hlet frawes woruleln hix scoting wisong Bareredpan . The " . Jron bargh was ady . be anand mive to chele 4s paas Ewers and in wanked ans Hict prorco 700 marly this copins rutize Ifend as pilint ) douns for Jish alver he bias unalalist eontatss elercor tlade conrat bath in quoster of thorcentist , treat for thelur and to .n = = Thate itter A F @-@ Iputurss occomanjottice . Nole ferg rited the Mall <unk> the gutureich ir . The . }
	\end{quote}
	
	One can see the spelling is mostly wrong except for short words such as \textit{with}, \textit{the} and \textit{<unk>}. Also the network cannot use punctuation correctly.

	\item Character-level LSTM with $m=1024$, single layer. The short-range mutual information is captured well. Intermediate-range mutual information is partially captured. No long-range power-law mutual information is captured. 
	\begin{quote}
		\small
		\textit{= = Association = =}
		
		\textit{ Paris migrate to <unk> hired by his favourites . In 1822 , it moved off and by 1990 . Piano 's themes in his wife has adult seedlane as being outdoority of visitors in the writers ; are Voyager in 2009 , she lived in Harbar 's classmates where him the motion of lanual piano and structure . During the 20th century , some of the fountain feeds a " star colour sight ; the role of the cultivate spielly moves due to <unk> of she <unk> at least 26 waters between the breeding course of the entrance in <unk> Brilia . The series is electronic and introduction of the penissential <unk> itself , without low range . }
	\end{quote}
	
	One can see the spelling is mostly correct. The network can correctly generate the section title by balancing \textit{=} signs, and form short phrases. However, the long sentence is not grammatical, and is not clear what is the central topic of the paragraph.

	\item Word-level LSTM with $m=2048$, single layer. Long-range power-law mutual information up to $\tau\lesssim 150$ is captured. 
	\begin{quote}
		\small
		\textit{= = = = Long = = = =}
		
		\textit{ Miami began the second quarter by 23 – 18 , in the eighth quarter . The team , <unk> heavenly , and Robert Lester were considered to be the first year after of the season . While the punt was slightly less modest , the Hokies began their first drive into seven plays , but also had three kickoffs . Despite <unk> , the Hokie offense proceeded in the game to <unk> , encountering effects needed to blow the ball off . A team arranged for 10 yards for the most intensely <unk> @-@ to @-@ back and field goal , while Virginia Tech received the ball to begin to win the game during one of the Hokie mathematical games are on the final series into the end of the game . play said to gain the first and also commenced the Virginia Tech pass . Sawtooth State praised its one @-@ yard crowd quarterfinal , which it singled out to make a kicking fourth down . Tech recovered the ball away from an early Odile to bring their first down – 20 – 3 in field goal . }
	\end{quote}
	
	Although not all sentences are grammatical, one can see the paragraph is around the topic of a football game. 
	
	\item Word-level Transformer with $m=512$, four layers. There is long-range power-law mutual information up to $\tau\lesssim 500$, although this decay is much slower than that of the training set. 
	\begin{quote}
		\small
		
		\textit{= = Release and success = = }
		
		\textit{" My New 500 " was Dick Walker Music Command successful <unk> Yankovic and Guitar airfields received the comic EAA on October 25 , 2013 . Set as a weaken to the song and to exclusively <unk> , Bennett called the album a " flamboyant @-@ with @-@ a @-@ regular , le equipped @-@ of @-@ <unk> claim that , out of ' a lot of people , ' from a like a group . ' " <unk> thoroughly from Somaliland wrote that the image was more reduced with a <unk> @-@ themed and changed . He enjoyed writing that the " Chun of I 'd have unlikely the technical enough of this holiday sets " . They classified the song as one of the fourth albums of the sequence , comparing it as one of the best songs in the history . }
		
		\textit{= = Legacy and recording music = = }
		
		\textit{" Even with earth " was the first song in which singer surrounds to " participation of Too " as well as a set . The song appeared with rebels in Chains and several American singles . It entered a second and final poor songs , a single album from ordination . Although featuring the album video list to create " easily healing ( ' burials ) " , a song that areas " to suppress the Mississippi I hope down album " was a portion of quite Ning 's " songs . " During the release of My ! Hill " , a omitted track stand from Alice in Chains , which aired in Chains pressures soon released by then Mughal rock films . It was shortened to rubber in as many <unk> solo : " Need of disaster BDS music " before overcome by McCartney produce a song . " }
	\end{quote}
	
	Although many sentences are ungrammatical, it is obvious both paragraphs are devoted to music. Also, the content of the paragraph corresponds with the section title very well. 
	
	\item 117M GPT-2. Power-law  mutual information is robust in the whole range investigated (up to $\tau\sim 1000$). 
	
	\begin{quote}
		\small		
		\textit{Figure 1 View largeDownload slide Memory and dementia of brain regions exposed to economic crisis or loss of intellectual function. Buraimas, Seravia and Nelson12 data on chronicity from individuals at high risk for chronic Alzheimer's disease52–60, and the myocardial rise are shown. Prudemazadeva et al.13 agreed with their authors for attributed cognitive defects in cml backups42 but, ignoring the possibility that frequent orosequencing resulted in nonsignificant variability, they suggested a mechanistic connection to the known disability of any covariates such as these but, assuming that the association with cognitive dysfunction would be fine decision making, we considered different such factors for assessing this issue by integrating the ageing vs the cumulative death rate from dementia, Graves' disease, Riordan syndrome, eccentric bradycardia, or multiple hereditary head trauma [eg, systematic new variants97 13–14 14 ] with older data. Following retroviral assays by task because residual repeats to CD4+CD30rs are rare or if interactions with acute and chronic diseases are common. Together, Stephenson et al.30 found that boom decrease in GAT signal while it is predictable suggested that both the decline in or lack of decline in GAT change all factors the aging ubiquitously but that everyone repeated the same comparison by repeated chronic disease (integral SAMS site databases). Bloomfield20 was also associated with mortality in our analyses but, data remaining already are inconclusive44 with the liability [BD35] being mild, age 59. Lack of CMV increase observed in aged people 54–61 in r10 or 6.33 microoriolvveg and suggestive that VSAP can be passed on from generations 1–6 with genome ascertainment. Cooper et al.10 suggested that CJ encounters were another connection to the body hyperrisk."}
	\end{quote}

	Although many sentences are ungrammatical, it is obvious the long paragraph is part of a a biomedical research paper. 
\end{itemize}

\section{An Exact Mapping Between Boltzmann Distribution and Conditional Distribution}
In this section, we show an exact example of sampling Boltzmann distribution $p(x_1,\ldots,x_t)=e^{-\beta H(x_1,\ldots,x_t)}$ conditionally according to 
\begin{equation}
p(x_1,\ldots,x_t)=\prod_{i=1}^t p(x_i|x_1,\ldots,x_{i-1}). \label{eqapp:prod}
\end{equation}

Consider the one-dimensional long-range Ising model with open boundary condition, whose Hamiltonian is given by
\begin{equation}
H=\sum_{i<j=1}^N J(|i-j|)s_is_j,
\end{equation}
where $s_i=\pm 1$ is the spin. At thermal equilibrium in the canonical ensemble, the probability of a spin configuration is given by the Boltzmann distribution:
\begin{equation}
p(s_1,\ldots,s_N)=\frac{1}{Z}e^{-\beta\sum_{i,j=1}^N J(|i-j|)s_is_j },
\end{equation}
where $\beta\equiv 1/k_BT$ is the inverse temperature and $Z\equiv \sum_{s_1,\ldots,s_N=\pm 1}$ is the partition function. The conditional distribution can be computed using Bayes' theorem:
\begin{align}
p(s_t|s_{t-1},\ldots,s_1)=&\frac{p(s_t,s_{t-1},\ldots,s_1)}{p(s_{t-1},\ldots,s_1)} \\
=&\frac{\sum_{s_{t+1},\ldots,s_N}e^{-\beta\sum_{i,j=1}^N J(|i-j|)s_is_j }}{\sum_{s_t,s_{t+1},\ldots,s_N}e^{-\beta\sum_{i,j=1}^N J(|i-j|)s_is_j }}
\end{align}
In the Hamiltonian, there are $N(N-1)/2$ pairs in the summation. We divide these pairs into the following groups:
\begin{enumerate}
	\item $\alpha=\{(i, j): i<j, 1\leq i,j\leq t-1\}$, i.e. interactions within the first $t-1$ spins;
	\item $\beta=\{(i, j): i<j, 1\leq i\leq t-1, t+1\leq j\leq N\}$, i.e. interactions between the first $t-1$ spins and the last $N-t$ spins;
	\item $\gamma=\{(i, j): i<j, t+1\leq i,j\leq N\}$, i.e. interactions within the last $N-t$ spins;
	\item $\delta=\{(i, t): 1\leq i\leq t-1\}$, i.e. interactions between the first $t-1$ spins and the $t$-th spin;
	\item $\epsilon=\{(t, j): t+1\leq j\leq N\}$, i.e. interactions between the last $N-t$ spins and the $t$-th spin.
\end{enumerate}

Because the summation does not involve pairs in group $\alpha$, group $\alpha$ from numerators and denominators simply cancel each other. Without further assumptions, the conditional distribution cannot be simplified further. The conditional probability is intractable, i.e. exponentially hard. 

We now assume the interaction is only nearest-neighbor
\begin{equation}
J(|i-j|)=\begin{cases}
J, & |i-j|=1, \\
0, & \text{otherwise}.
\end{cases}
\end{equation}
Under this assumption, group $\beta$ is empty. It follows
\begin{equation}
p(s_t|s_{t-1},\ldots,s_1)=\frac{e^{-\beta J s_{t-1}s_t}\sum_{s_{t+1},\ldots,s_N}e^{-\beta J s_ts_{t+1} } e^{-\beta J \sum_{i=t+1}^{N-1} s_is_{i+1} }}{\sum_{s_t}e^{-\beta J s_{t-1}s_t}\sum_{s_{t+1},\ldots,s_N}e^{-\beta Js_ts_{t+1} }e^{-\beta J \sum_{i=t+1}^{N-1} s_is_{i+1} }}.
\end{equation}
The factor $ f(s_t)=\sum_{s_{t+1},\ldots,s_N}e^{-\beta J s_ts_{t+1} } e^{-\beta J \sum_{i=t+1}^{N-1} s_is_{i+1} }$ is actually an $s_t$-independent constant because of the $\mathbb{Z}_2$ spin flip symmetry of the Ising model. To see this, simply let $s_i\to -s_i$  for $t+1\leq i\leq N$. Therefore, the factor cancels from the denominator and the numerator. We end up having
\begin{equation}
p(s_t|s_{t-1},\ldots,s_1)=\frac{e^{-\beta J s_{t-1}s_t}}{\sum_{s_t}e^{-\beta J s_{t-1}s_t}}=\frac{e^{-\beta J s_{t-1}s_t}}{2\cosh\left(\beta J s_{t-1}\right)}, \label{eqapp:ising}
\end{equation}
which only depends on $s_{t-1}$. 

To summarize, in nearest-neighbor Ising model, the Boltzmann distribution can be sampled conditionally according to Equations \eqref{eqapp:prod} and \eqref{eqapp:ising}. Note that in this way, Ising model is essentially a one-gram model with two symbols. Therefore, the mutual information must decay exponentially. This result is consistent with van-Hove's theorem in statistical mechanics, which states that one-dimensional models with short-range interactions cannot be critical at finite temperature. 

Note that we exploited two critical properties of Ising model in order to obtain the final result:
\begin{enumerate}
	\item The interaction is nearest-neighbor;
	\item The interaction has $\mathbb{Z}_2$ spin flip symmetry. 
\end{enumerate}
Relaxing any of them makes the calculation intractable.

\end{document}